\documentclass[preprint,12pt]{elsarticle}

\usepackage{lineno,hyperref}
\usepackage{amsmath}
\usepackage{color}
\usepackage{subcaption}
\usepackage{graphicx}
\usepackage{caption}
\usepackage{float}
\usepackage[labelfont=bf]{caption}
\usepackage{subcaption}
\usepackage{epstopdf}
\usepackage{enumitem}
\usepackage[normalem]{ulem}
\usepackage{color}
\usepackage{amssymb}

\newcommand\bX{\boldsymbol X}

\newcommand\bx{\boldsymbol x}

\newcommand\bU{\boldsymbol U}
\newcommand\bv{\boldsymbol v}

\newcommand\bT{\boldsymbol T}

\newcommand\bPhi{\boldsymbol{\Phi}}
\newcommand\bSigma{\boldsymbol{\Sigma}}

\definecolor{bleudefrance}{rgb}{0.19, 0.55, 0.91}

\modulolinenumbers[5]

\journal{Preprint}

\begin{document}

\begin{frontmatter}

\title{A predictive physics-aware hybrid reduced order model for reacting flows}

%% Group authors per affiliation:
\author{A. Corrochano$^{1,*}$ , R.S.M. Freitas$^{2,3}$,\\ A. Parente$^{2,3}$ and S. Le Clainche$^1$}
\cortext[cor1]{Corresponding author\\
E-mail addresses: adrian.corrochanoc@upm.es (A. Corrochano), rodolfo.da.silva.machado.de.freit@ulb.be (R.S.M. Freitas), Alessandro.Parente@ulb.be (A. Parente), soledad.leclainche@upm.es (S. Le Clainche).} 

\address{$^1$ E.T.S.I. Aeron\'autica y del Espacio,%\\
Universidad Polit\'ecnica de Madrid, Spain}%\\
\address{$^2$ Universit\'{e} Libre de Bruxelles, \'{E}cole polytechnique de Bruxelles, Aero-Thermo-Mechanics Laboratory, Brussels, Belgium }
\address{$^3$ Université Libre de Bruxelles and Vrije Universiteit Brussel, Brussels Institute for Thermal-Fluid Systems and Clean Energy (BRITE), Brussels, Belgium}

\begin{abstract}

In this work, a new hybrid predictive Reduced Order Model (ROM) is proposed to solve reacting flow problems. This algorithm is based on a dimensionality reduction using Proper Orthogonal Decomposition (POD) combined with deep learning architectures. The number of degrees of freedom is reduced from thousands of temporal points to a few POD modes with their corresponding temporal coefficients. Two different deep learning architectures have been tested to predict the temporal coefficients, based on recursive (RNN) and convolutional (CNN) neural networks. From each architecture, different models have been created to understand the behavior of each parameter of the neural network. Results show that these architectures are able to predict the temporal coefficients of the POD modes, as well as the whole snapshots. The RNN shows lower prediction error for all the variables analyzed. The model was also found capable of predicting more complex simulations showing transfer learning capabilities. 

\end{abstract}


\end{frontmatter}

%\linenumbers

\section{Introduction}

The development of combustion systems will continue to play a role in the future mainly for those applications that require high energy density, thus motivating fundamental research aiming to develop fuel-flexible, efficient, and clean combustion technologies. Consequently, strong efforts have been dedicated and will continue to be carried out by the combustion community to understand the underlying physics of reacting flows, motivated by the pressing need to leverage the performance and the efficiency of the combustion technologies and with the aim of finding new fuel-flexible alternatives to the current use of fossil fuels.

To address these challenges, computational models can advance the current understanding of reacting flows. However, the large number of species involved in combustion processes (curse of dimensionality), the wide variety of spatial and time scales, and the non-linear turbulence-chemistry interactions limit the use of numerical simulation tools in the design and operation of large scale combustion systems~\cite{doi:10.1098/rsta.2002.0990}. Reduced order models (ROMs) are an attractive solution to overcome such limitations, as they can act as proxies for high-fidelity models~\cite{parente2011investigation, bellemans2018feature, Coussement2013}.

With the advent of data science~\cite{IHME2022101010,Zdybal_VKI_2022,ML_aircraft_review}, new challenges and research opportunities are emerging, related to how to make the best use of the data provided by experiments and high-fidelity simulations. A thorough description of the use of data-driven machine learning in combustion can be found in Ref. \cite{Zhou2022}. Two main models of machine learning ROMs have been used in combustion: unsupervised and supervised learning. Two algorithms that have been used to create unsupervised ROMs in combustion systems are Principal Component Analysis (PCA)\cite{Jolliffe2011} and Dynamic Mode Decomposition (DMD) \cite{Schmid10}. PCA is an algorithm that decomposes the data into principal components, which are a linear combination of the original variables. With this method, it is possible to project the original data into a lower-dimensional manifold. It has been used to re-parametrize the thermo-chemical state, thus speeding up computational fluid dynamics (CFD) simulations \cite{isaac2014reduced,bellemans2017reduced}.
PCA has also been used for clustering \cite{d2020adaptive, d2020impact}, feature extraction and selection \cite{d2020analysis, d2021feature, bellemans2018feature, krzanowski1987selection} and data analysis\cite{d2020analysis,d2020unsupervised}.

DMD is a data-driven technique proposed for the analysis of unsteady systems. It can be used for dimensionality reduction or to study flow patterns in combustion, related to flow instabilities \cite{Souvick2013, Quinlan2014, Huang2016, Motheau2014}. More recently, Higher Order Dynamic Mode Decomposition (HODMD) \cite{LeClaincheVega17} was found to be more robust for the analysis of complex flows \cite{Corrochano, CorrochanoJAE, Munoz2022, Lazpita22}. This algorithm was also validated for the analysis of reacting flows \cite{Corrochano22}.

In the field of supervised learning, many studies have been carried out to create ROMs. Recently, advances in ROMs for fluid dynamics were reviewed by Brunton et al.~\cite{Brunton_ML_fluids}. They highlight the capabilities of physics-aware machine learning to improve fluid simulations. Also, different machine learning models have been implemented to leverage combustion simulations in several different applications. Huang et al. \cite{Huang20} used a convolutional neural network (CNN) to reconstruct a 3D flame, based on 2D reconstruction. Li et al. \cite{Li20} used a single hidden-layer neural network (NN) for accelerating the design of new clean fuels. Some models have been used to optimize engine performance and control, as in Refs. \cite{Wong2010, Badra20}. Sharma et al. \cite{Sharma20} constructed an artificial neural network to model a hydrogen combustion process. Nikitin et al. \cite{Nikitin22} proposed a model for hydrogen oxidation, training a neural network able to predict in time with several initial conditions of the system.

The combination of ROMs and deep learning represents an attractive opportunity to enhance combustion simulations. In this way, Zhang et al. \cite{ZHANG2020156} proposed a ROM combining proper orthogonal decomposition (POD)~\cite{Lumley67} and NNs for the reconstruction of the cellular surface of gaseous detonation wave surface based on a post-surface flow field. Wang et al.~\cite{WANG2019289} proposed an approach combining POD and NNs providing reliable solutions for the quasi-one dimensional Continuously Variable Resonance Combustor. In the present work, a hybrid ROM is proposed by combining POD for dimensionality reduction with deep learning architectures based on physical principles. More specifically, POD reduces the data to an expansion of spatial modes, orthogonal in space, and temporal modes. This decomposition is then fed to two deep learning algorithms to discover the underlying physical dynamics of the system using the embedded information in the temporal coefficients. The combination of these two techniques has been used in fluid mechanics, either to reconstruct a flow field \cite{Freitag2018, Guemes2019} or to construct predictive models \cite{Abadia2022}. This article leverages the work presented by Abadía-Heredia et al. \cite{Abadia2022}, expanding the applicability of the predictive ROM, through the combination of POD with neural network architectures. 

The current work is organized as follows. In Section \ref{sec:meth}, the hybrid ROM algorithm is explained. In Section \ref{sec:Sim}, the numerical simulation and the database extracted is briefly explained. The main results and conclusions are presented in Sections \ref{sec:results} and \ref{sec:conclusions}, respectively.

%------------------------------------------------------------------
\section{Computational framework \label{sec:meth}}

The algorithm used in the present work is a combination of pre-processing techniques, widely used in combustion, POD for dimensionality reduction, and deep learning predictive models based on recursive (RNN) and convolutional (CNN) neural networks.

Firstly, the data is organized into a snapshot matrix as follows
\begin{equation}
	\boldsymbol{X} = [\bv_1,\bv_2,\ldots,\bv_k,\bv_{k+1},\ldots,\bv_{K-1},\bv_K], 
	\label{ab0}
\end{equation}
\noindent where $\boldsymbol{X} \in \mathbb{R}^{J \times K}$ and $\bv_k$ is the state variable of the reacting flow at a time instant $t_k$. Here, $K$ is the number of snapshots, equidistant in time, while $J = N_v \times N_x\times N_y \times N_z$ being $N_v$ the number of variables, and $N_x$, $N_y$ and $N_z$ the number of grid points in the streamwise, normal and spanwise spatial components of the domain. The application presented in this article is a two-dimensional, axisymmetric case, so the dimension of the snapshot matrix Eq.~\eqref{ab0} is ${(N_v \cdot N_x \cdot N_y) \times K}$.

\subsection{Pre-processing techniques\label{sec:centering}}

Since combustion is intrinsically a multi-scale problem, with the thermodynamics states describing the whole system differing by several orders of magnitudes, the database must be processed to obtain meaningful results. 
Before introducing the variables into the snapshot matrix $\boldsymbol{X}$, each one has to be pre-processed. The techniques used in this study are centering and scaling.
Centering means the subtraction of the temporal mean of the variable, so the analysis is focused on the fluctuations around the mean. 
Scaling is motivated by the necessity of comparing variables on the same basis. In this analysis, the variables have been scaled with their standard deviation $\sigma_j$. This scaling method, called \textit{auto scaling}, gives all the variables the same importance \cite{PCA}. The centering and scaling can be summarised in the same equation as
\begin{equation}
	\tilde{\bx}_j(t_k) = \dfrac{\bx_j(t_k)-\bar{x}_j}{\sigma_j}\label{e24},
\end{equation}
where $\bx_j$ is the \textit{j-th} variable, $\bar{x}_j$ the temporal mean and $\tilde{\bx}_j$ the scaled variable. 

\subsection{Proper Orthogonal Decomposition}

Once the database has been pre-processed, POD \cite{Lumley67} is applied to the database to reduce the dimensionality of the problem. The method decomposes the original data as a combination of POD modes $\bPhi_j(x,y)$, which are orthogonal in space and optimal in terms of preserved energy. The optimal modes are found by minimizing the mean square error between the POD approximation and the full-order model. Finally, this method models the flow field as a linear combination of POD modes with the temporal coefficients $c_j(t)$,
\begin{equation}
	\bX \simeq \sum_j c_j(t)\bPhi_j(x,y).
	\label{ab2}
\end{equation}
In this work, we employ the singular value decomposition (SVD) \cite{Sirovich87} algorithm to perform POD. This algorithm decomposes the original snapshot matrix into the POD modes $\bU \in \mathbb{R}^{J \times N}$, the temporal coefficients $\bT$ and the singular values $\bSigma$, which contains the amount of energy of each mode and its contribution to the reacting field, as
\begin{equation}
	\bX\simeq\bU\,\bSigma\,\bT^\top,\label{ab20}
\end{equation}
where $()^\top$ denotes the transpose matrix. The diagonal matrix $\bSigma$ contains the singular values $\sigma_1,\cdots,\sigma_N$, related with the $N$ selected POD modes. The modes are ranked in decreasing order by the singular values, where the most energetic modes (highest singular value) provide reliable information on the thermodynamics states dynamics. The number of POD modes $N$ can be set for a given fraction of preserved energy
\begin{equation}
	E(\%) = \frac{\sum_{j = 1}^{N} \sigma_j}{\sum_{j = 1}^{K} \sigma_j}.\label{ab21}
\end{equation}
Selecting a subset of POD modes eliminates the noise level in experiments or the small scales of the flow structures, retaining just the largest ones, which are connected to the coherent structures of the flow leading to the main dynamics.

Finally, the accuracy of SVD considering N modes to approximate the original data can be measured by means of the relative root mean square error (RRMSE) as
\begin{equation}
	RRMSE^{POD}=\frac{||{\bX}-\bU\,\bSigma\,\bT^\top||}{||{\bX}||},\label{RRMSEpod}
\end{equation}
where $||\cdot||$ is the L2-norm.

Furthermore, we can define $\hat{\bT}=\bSigma \bT^T$ and the reacting field can be written as,

\begin{equation}
	\bX\simeq\bU\,\hat{\bT},
\end{equation}

\noindent where $\hat{\bT} \in \mathbb{R}^{N\times K}$ is the matrix containing temporal modes. Specifically, the temporal modes contain the underlying dynamics of the reacting flow field. Consequently, each row of the temporal modes matrix corresponds to the N temporal coefficients $c_j(t)$. Knowing the temporal coefficients from time step $1$ to $K$, it is possible to predict (extrapolate) the next $K^{*}$ time steps, obtaining a new matrix $\hat{\bT}^{*}$ of dimensions $N \times K^{*}$. Hence, it is possible to reconstruct the original thermodynamics states with the POD modes obtaining a new snapshot matrix as  

\begin{equation}
	\hat{\bX^*} = \bU \hat{\bT}^{*} \label{eq:8},
\end{equation}

\noindent where $K+1$, $K+2$, $\cdots$, $K^{*}$ are the snapshots of the reacting flow predictions. Here, deep learning techniques are used as a time integrator to advance temporal modes in time. This is usually done  with Galerkin projection and solution of ODEs for the coefficients, with a higher computational cost \cite{Rapun2010, LeClainche201779}. Such models allow the introduction of physics-guided information in the optimization process, making them physics aware and able to extrapolate for unseen snapshots of the thermodynamic states.

\subsection{Deep learning predictive models \label{sec:DLmodels}}

This section introduces the two deep learning models that have been used to predict the temporal coefficients $ \hat{\bT}$. The idea behind the prediction is to select an initial sequence of $q$ snapshots to predict the snapshot at time $t+1$, where each snapshot corresponds to a column in the temporal matrix $\hat{\bT}$. More specifically, the temporal coefficients $\hat{\bT}_{t+1}$ are predicted using the $q$ previous snapshots given by $\hat{\bT}_t, \hat{\bT}_{t-1}, \cdots, \hat{\bT}_{t-q+1}$. These new snapshots are the thermodynamic state predictions, the essence of the ROM. The deep learning architectures use the information of the previous $q$ time steps to predict the next $p$ time steps.

In the present work, two different deep learning models are used to act as time integrators to solve the temporal coefficients. Firstly, a recurrent model is constructed by combining long short-term memory (LSTM)~\cite{LSTM} layers and fully connected (FC) layers. Recurrent models learn the dynamics of the system by introducing a feedback mechanism in the hidden layers, allowing the storage of the states (also called memory) of previous inputs to generate the next output of the sequence. Specifically, LSTM networks have been applied with success to reproduce the temporal dynamics of turbulent flows~\cite{BORRELLI2022109010}, and here we aim to explore the potential of such a model in predicting the temporal dynamics of the reacting flows. 

On the other hand, a convolutional model is built composed of one-dimensional convolutional layers (Conv1D) followed by FC layers. CNNs have  great potential to discover repeated patterns in the time series using convolutional filters and extract intrinsic features directly from data without prior knowledge~\cite{CNN_tseries1, CNN_tseries2, CNN_tseries3}. 

In the present work, these models are chosen due to their robustness and ability to generalize for future time instants not used in the training. Furthermore, both models are characterized by a relatively small number of parameters to be learned, which is useful to avoid overfitting. The architecture details of both models are inspired by Ref. \cite{Abadia2022} and presented in Tab. \ref{tab:ML1} and \ref{tab:ML2}. The codes to construct the models are available from Ref. \cite{NNcodes}.
\begin{table}[H] 
	\centering
	\scalebox{0.8}{
		\begin{tabular}{|c|c|c|c|c|}
			\hline
			\# Layer & Layer details & \# Neurons & Activation Function& Dimension\\\hline
			$0$ & Input & $N$ &  & $10\times N$\\
			$1$ & LSTM & $100$ & ReLU$^1$ / ELU$^2$ & $100$\\
			$2$ & FC & $6 * 100$ & ReLU$^1$ / ELU$^2$ & $6 * 100$ \\
			$3$ & Reshape &  &  & $6 \times 100$ \\
			$4$ & FC & $80$ & ReLU$^1$ / ELU$^2$ & $6 \times 80$ \\
			$5$ & Split($p$) & & & $80$ \\
			$6$ & $p * $ FC & $N$ & Sigmoid$^1$ / Tanh$^2$ & $N$ \\\hline
		\end{tabular}
	}
\caption{Architecture details in the RNN. The architecture is the same as in Ref. \cite{NNcodes}. $^1$ Activation function used in the original model. $^2$ Activation function proposed in the present work, which will be further explained.\label{tab:ML1}}
\end{table}
\begin{table}[H] 
	\centering
	\scalebox{0.8}{
		\begin{tabular}{|c|c|c|c|c|c|c|c|}
			\hline
			\# Layer & Layer details & \# Neurons & Kernel size & Stride& Padding& Activation& Dimension\\\hline
			$0$& Input & $N$ & & & & & $10\times N$\\
			$1$& Conv 1D & $30$ & $3$& $1$& No & ReLU$^1$ / ELU$^2$ & $8\times 30$\\
			$2$& Conv 1D & $60$ & $3$& $1$& No & ReLU$^1$ / ELU$^2$ & $6\times 60$\\
			$3$ & Flatten & & & & & & $360$\\
			$4$& FC & $100$ & & & & ReLU$^1$ / ELU$^2$ & $100$ \\
			$5$& FC & $100$ & & & & ReLU$^1$ / ELU$^2$ & $100$ \\
			$6$ & Split($p$) & & & & & & $100$ \\
			$7$& $p * $ FC & $N$ & & & & Sigmoid$^1$ / Tanh$^2$ & $N$ \\\hline
	\end{tabular}}
	\caption{Architecture details in the CNN. $^1$ Activation function used in the original model. $^2$ Activation function proposed in the present work, which will be further explained. \label{tab:ML2}}
\end{table}
To train the machine learning models, a mean-squared error (MSE) loss function between the true $\hat{\bT}$ and predicted $\hat{\bT}^*$ temporal modes at each time instant $t$ is used, 

\begin{equation}
    MSE = \frac{1}{N_{K}}\sum_{t=1}^{N_K} \left( \frac{1}{N} \|\hat{\bT}_{t} - \hat{\bT}_{t}^{*} \|^2 \right),
\end{equation}

\noindent  where $N$ is the number of singular values and $N_K$ is the number of snapshots used during the training process. Further, we propose a physics-aware mean-square error (PA-MSE) loss function which constrains the learning such as the mass balance is conserved in each time instant, given as

\begin{equation}\label{eq:custom}
    PA-MSE = \frac{1}{N_{K}}\sum_{t=1}^{N_K} \left( \frac{1}{N} \|\hat{\bT}_{t} - \hat{\bT}_{t}^{*} \|^2 \right) + \frac{1}{N_{K}}\sum_{t=1}^{N_K} \left( \|\sum_{s=1}^{N_s}Y_s(t) - \sum_{s=1}^{N_s}Y_s^{*}(t) \|^2 \right), 
\end{equation}

\noindent where $N_s$ is the number of chemical species in the reacting flow. Moreover, $Y_s$ and $Y_s^{*}$ are the mass fractions of species reconstructed from the true $\hat{\bT}$ and predicted $\hat{\bT}^*$ temporal modes, respectively. It is worth remarking that the same importance is given to both components in the loss function, \emph{i.e.}, PA-MSE was not weighted. Specifically, finding the hyperparameters which return a good balance between the loss components is still a topic of research in the ML community. The weights may act as hyperparameters in the training process, and further research efforts should be dedicated to searching these hyperparameters.

\begin{figure}
	\centering
	\includegraphics[width=0.95\textwidth]{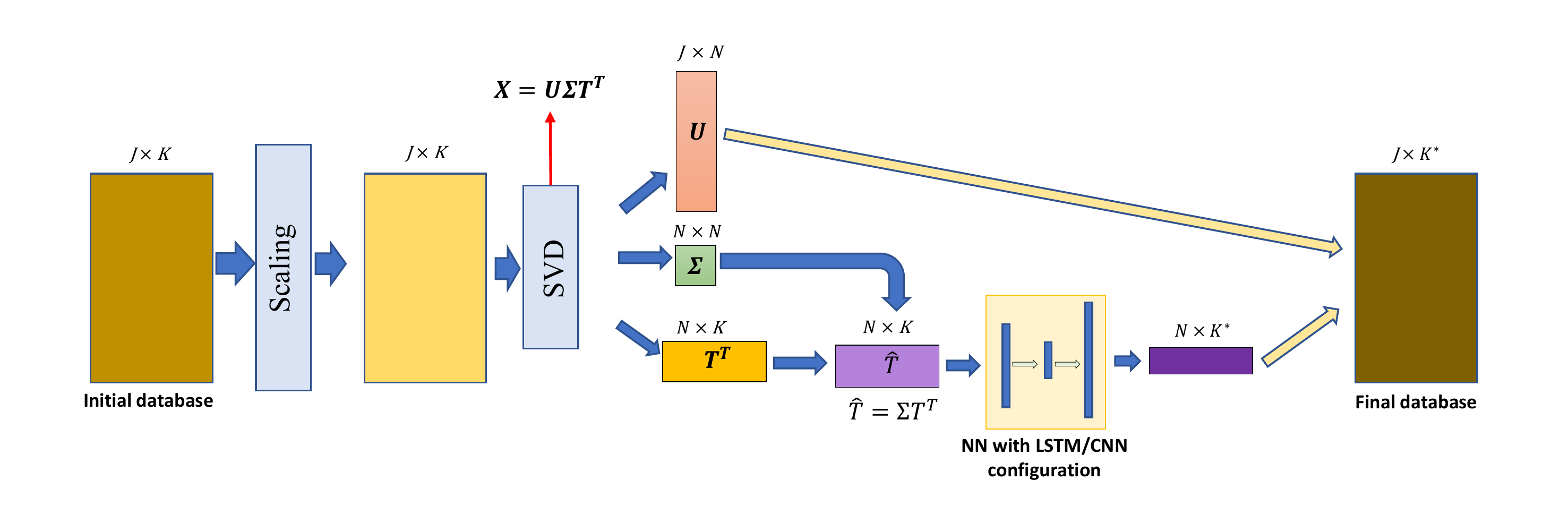}\\
	\caption{Sketch containing the main steps of the implemented algorithm. The initial database contains the original data and the final database is calculated by multiplying the POD modes and the predicted temporal coefficients, as in Eq. \ref{eq:8}}
	\label{fig:Figure1}
\end{figure}

An overview of the whole computational framework is summarized in Fig.~\ref{fig:Figure1}.

\section{Numerical simulation \label{sec:Sim}}

This section briefly introduces the numerical simulation of the reactive flow and the extraction of the database used in the study. The considered numerical simulation is representative of an axisymmetric, time-varying, laminar co-flow flame, where the fuel is nitrogen-diluted methane (65\% methane, 35\% nitrogen, on a molar basis) and the oxidizer is air. The oxidizer is injected at a constant velocity of $35$ cm/s and the fuel is injected with a parabolic profile with a perturbation in time $t$, as

\begin{equation}
\label{eq:parabolic_profile}
	v(r,t) = v_{max} \left( 1 - \frac{r^2}{R^2} \right) [1 + A \ sin(2 \pi f \ t)],
\end{equation}

\noindent where $v_{max} = 70$ cm/s is the maximum velocity, $r$ is the radial coordinate, $R$ the internal radius of the nozzle, $A = 0.25$ is the amplitude of the perturbation and $f = 20 Hz$ is the frequency of the perturbation. For the numerical simulation, \texttt{GRI-Mech\ 3.0} \cite{smith1999gri} was employed. It consists of 325 elementary reactions containing 53 species with C1-C2 hydrocarbons. The CFD simulation was carried out by using \texttt{LaminarSMOKE}, an \texttt{OpenFOAM}-based operator-splitting solver by Cuoci \emph{et al.} \cite{cuoci2013numerical}. More information about the numerical settings can be found in Refs. \cite{d2020adaptive,d2020impact}. 

A database was extracted consisting of temperature and the $9$ chemical species with the highest maximum concentration, which makes the total number of variables equal to $N_v = 10$. These chemical species are:

\begin{itemize}
	\item Air components: $O_2$ and $N_2$
	\item Fuel components: $CH_4$
	\item Main oxidation products: $CO_2$ and $H_2O$
	\item Minor species: $C_2H_2$, $C_2H_4$, $CO$ and $OH$
\end{itemize}

\begin{figure}
	\centering
	\includegraphics[height=0.5\textwidth]{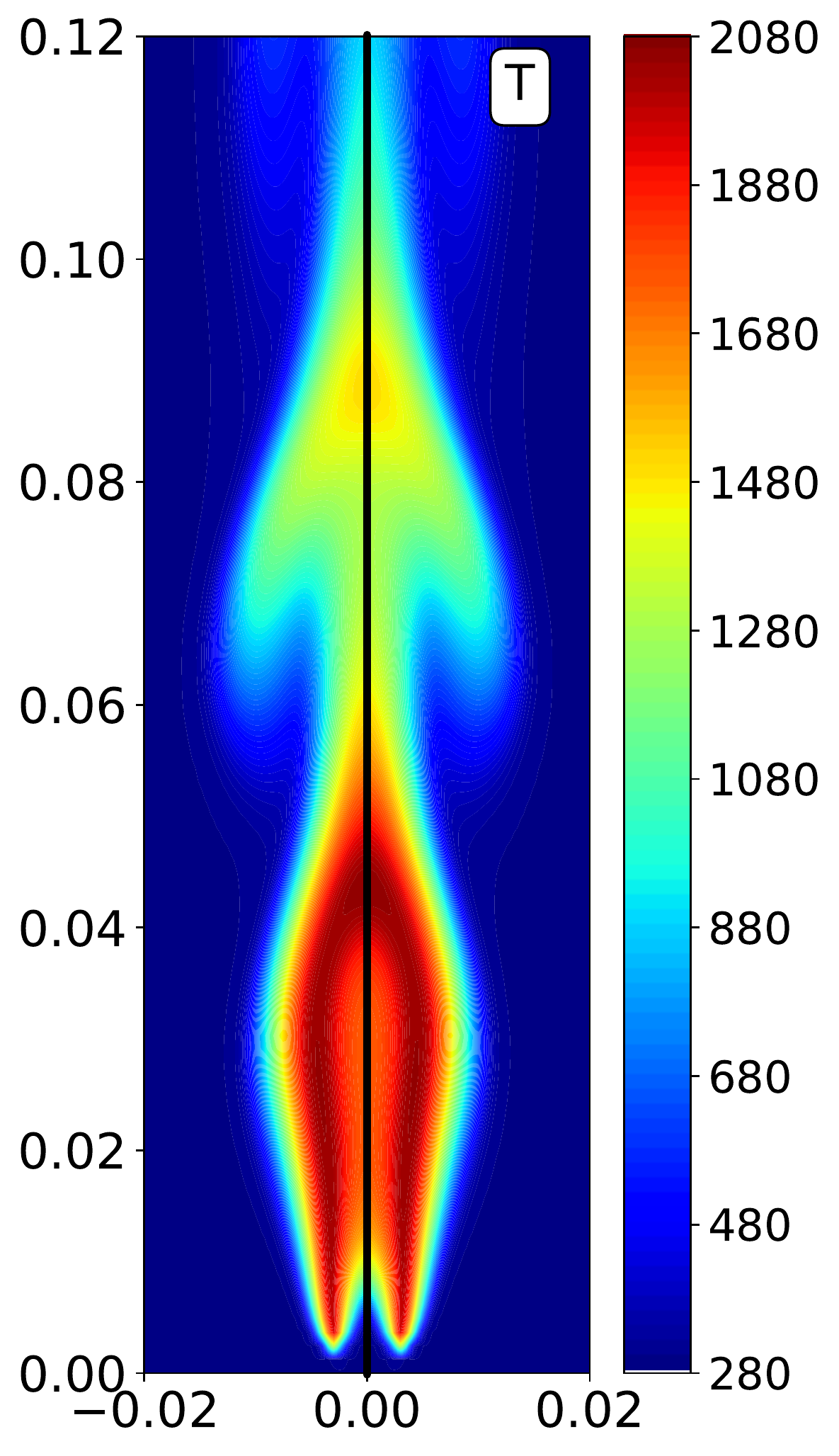}
	\includegraphics[height=0.5\textwidth]{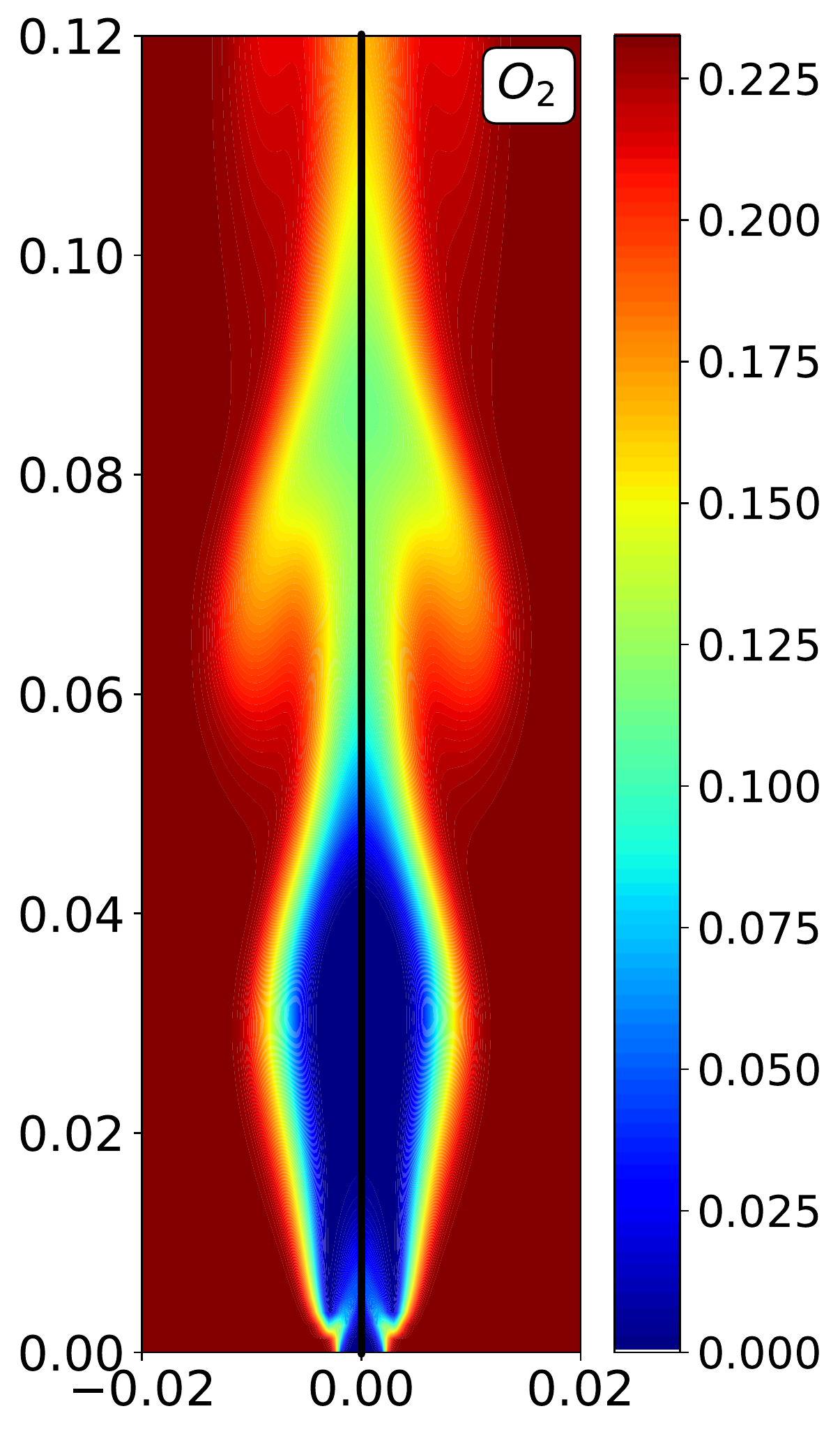}
	\includegraphics[height=0.5\textwidth]{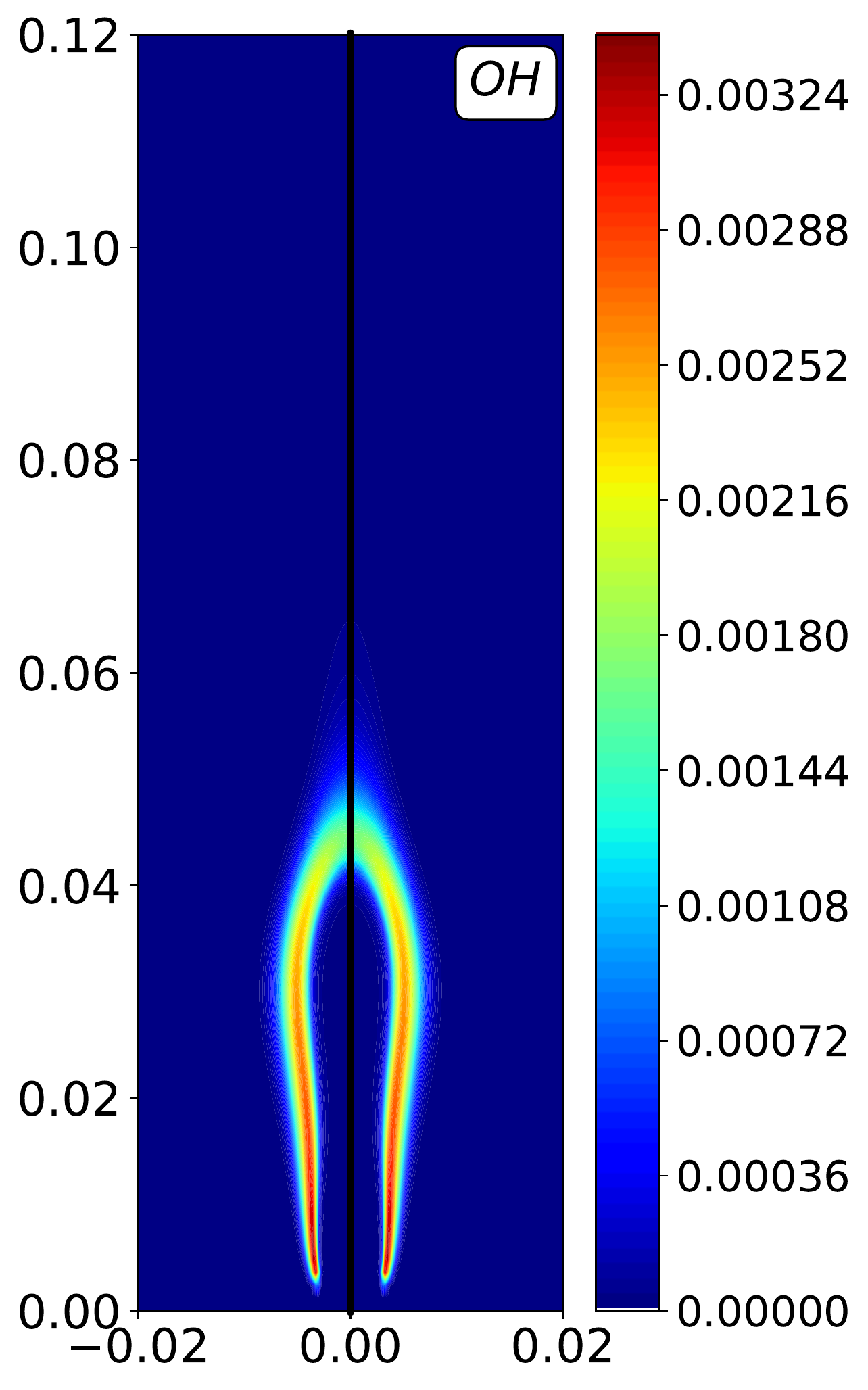}\\
	\caption{Instantaneous image of the temperature, oxygen, and OH mass fractions (from left to right). The black line represents the axi-symmetric axis.}
	\label{fig:Figure2}
\end{figure}

The number of snapshots extracted is $n_t = 999$, equidistant in time with $\Delta t = 2.5 \times 10^{-4} s$. Hence, the time interval covered by the dataset is $T = n_t \times \Delta t \simeq 0.25s$, approximately $5$ cycles of the perturbation of the velocity profile. Finally, the database has been extracted in a structured mesh of dimensions $100 \times 75$, in the streamwise and normal directions, respectively. In tensor form, the dimensions of the database are $N_v \times N_x \times N_y \times n_t = 10 \times 100 \times 75 \times 999$. In Fig. \ref{fig:Figure2}, a representative snapshot of the database has been plotted.

\section{Results and Discussions \label{sec:results}}

In this section, the ability of the proposed deep learning models to predict the temporal modes is evaluated. The database is first reshaped from a fourth-dimensional tensor ($N_v \times N_x \times N_y \times n_t$) to a two-dimensional matrix ($N_v \cdot N_x \cdot N_y \times n_t$). SVD is applied to the matrix and the first $18$ SVD modes are taken. These modes capture $80\%$ of the total energy and contain the largest scales of the simulation, giving a reconstruction error (RRMSE)  of $0.022$. The reconstruction error of each variable can be found in Tab. \ref{tab:RRMSE_trunc}.
\begin{table}[H] 
	\centering
    \begin{tabular}{|c|c|c|c|c|}
    \hline
     $T$ & $C_2H_2$ & $C_2H_4$ & $CH_4$ & $CO$ \\ 
     \hline
     $0.022$ & $0.101$ & $0.08$ & $0.024$ & $0.061$ \\
     \hline \hline
     $CO_2$ & $H_2O$& $N_2$ & $O_2$ & $OH$  \\
     \hline
     $0.03$ & $0.02$ & $0.0007$ & $0.009$ & $0.0146$ \\
     \hline
    \end{tabular}
    \caption{Reconstruction error of each variable studied.}
    \label{tab:RRMSE_trunc}
\end{table}

The database has been separated into three sequential blocks, which are used for the training, validation, and testing of the deep learning models. The test set includes the last $20\%$ of the snapshots. From the first $80\%$ of the snapshots, the training set and validation sets include the first $85\%$ and the last $15\%$ of the snapshots, respectively. Hence the total number of time steps is given by $K_{training} +K_{validation} + K_{test} = n_t$. Taking into account the nomenclature in Fig. \ref{fig:Figure1}, $K_{training}+K_{validation} = K$ and $K_{test} = K^{*}$. Thus, the matrix for the deep learning models has dimensions  $N \times K = 18 \times 800$, much smaller than the original one, reducing the computational cost by a large extent.

In the training process, each column of the temporal modes $\hat{\bT}$ ($\bv_j$) is scaled with the sum of the maximum values of all the columns, as

\begin{equation}
	\hat{\bv_j} = \dfrac{\bv_j}{\sum_{j = 1}^{N}\text{max}|\bv_j|}\label{eq:Kapta}.
\end{equation}

\noindent That scaling ranges the values of $\hat{\bT}$ between -0.15 and 0.15. Thus, the activation function of the output layer in the original model was changed from a sigmoid function, which outputs values between $0$ and $1$, to a hyperbolic tangent function, which constrains the outputs in the range $[-1,1]$. 

The deep learning models use the information of $q = 10$ previous snapshots to predict the next $p = 6$ time-ahead snapshots, as in Ref.~\cite{LopezMartinetal21}. \textit{Early stopping} was used to obtain robust model parameters: the training stops when the MSE is not reduced after a certain number of epochs (patience period).
Adam optimizer \cite{Kingmaetal} is used with default values for the parameters (learning rate $\alpha = 0.001$, $\beta_1 = 0.9$, $\beta_2 = 0.999$ and $\epsilon = 10^{-8}$). Mini batch gradient descent was used with a batch size equal to $12$ and the models were trained over $100$ epochs and a patience period of $20$ epochs. Moreover, the learning rate was set to reduce when a metric was observed to stop improving. This reduction is made by multiplying the learning rate by a factor smaller than 1 after a fixed number of epochs.

The accuracy of the algorithm is measured using the distance between the true $\hat{\bT}$ and predicted $\hat{\bT}^{*}$ temporal modes. To evaluate the models quality, we consider the RRMSE, calculated on the predicted snapshots of each variable, as

\begin{equation}
	RRMSE_j^{NN}=\frac{\|\tilde{\bx_j}^\text{predicted}-\tilde{\bx_j}\|}{\|\tilde{\bx_j}\|},
\end{equation}

\noindent where $\tilde{\bx_j}$ is the $j$-th variable and $\tilde{\bx_j}^\text{predicted}$ its prediction.

In subsection \ref{sec:Hyper}, the selection of the hyperparameters on the LSTM and CNN is analyzed to search for a robust scheme, starting from the original models  \cite{Abadia2022} up to the ones with the lowest $RRMSE$ for all the variables analyzed. In subsection \ref{sec:com}, the models with the best hyperparameters are compared to study which one, LSTM and CNN, is more suitable for the time predictions.

\subsection{Influence of hyper parameters \label{sec:Hyper}}

The influence of the key hyperparameters of deep learning models is first investigated. The models were optimized for the reacting flow problem using trial-and-error to search for key hyperparameters that give the lowest reconstruction error (RRMSE). The way of setting the parameters to ensure the best performance in NN architectures is still generally unknown. We divide the training and validation process into six different cases, combining the modifications from Tab.~\ref{tab:changes} to analyze the influence of the different hyperparameters.

\begin{figure}
	\centering
	\includegraphics[width=0.95\textwidth]{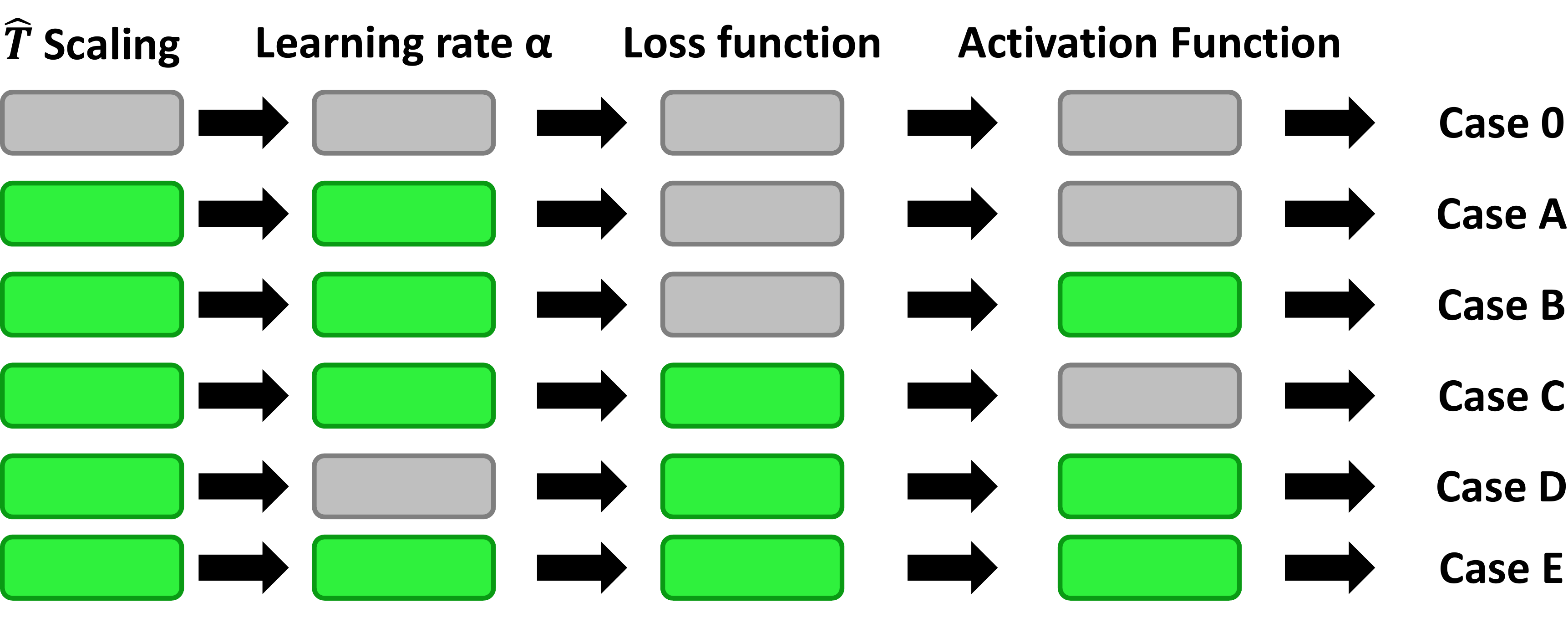}\\
	\caption{Sketch containing the cases investigated to assess the influence of the hyperparameters.
	If no changes with respect to the original model are observed, the square is coloured grey. In case of a change, the square is coloured green. Changes are listed on Tab. \ref{tab:changes}.}
	\label{fig:FigureC}
\end{figure}

\begin{table}[H] 
	\centering
	\scalebox{0.8}{
		\begin{tabular}{|c|c|c|}
			\hline
			 & Original NN & Changes \\ 
			 \hline
			$\hat{\bT}$ Scaling & Range scaling & \textit{Kaptanoglu et al.} \cite{Kaptanoglu21} Eq.\eqref{eq:Kapta}\\
			\hline
			Learning rate $\alpha$ & $0.001$ & $0.005 (*0.8 \; \text{every 10 epochs})$\\
			\hline
			Loss function & MSE & PA-MSE, Eq.~\eqref{eq:custom} \\
			\hline
			\begin{tabular}{@{}c@{}}Activation function \\ of the hidden layers\end{tabular} & ReLU &  ELU \\
			\hline
		\end{tabular}
	}
	\caption{Changed proposed in the neural networks. The activation function of the output layer has been changed from {\it sigmoid} to {\it tanh} in all the cases with a change on the $\hat{\bT}$ Scaling. \label{tab:changes}}
\end{table}
The six cases proposed in this study are summarised in Fig. \ref{fig:FigureC}. In each case, changes have been made in some of the hyperparameters to test their influence on the performance of the neural network.

\begin{figure}
	\centering
	\includegraphics[height=0.3\textwidth]{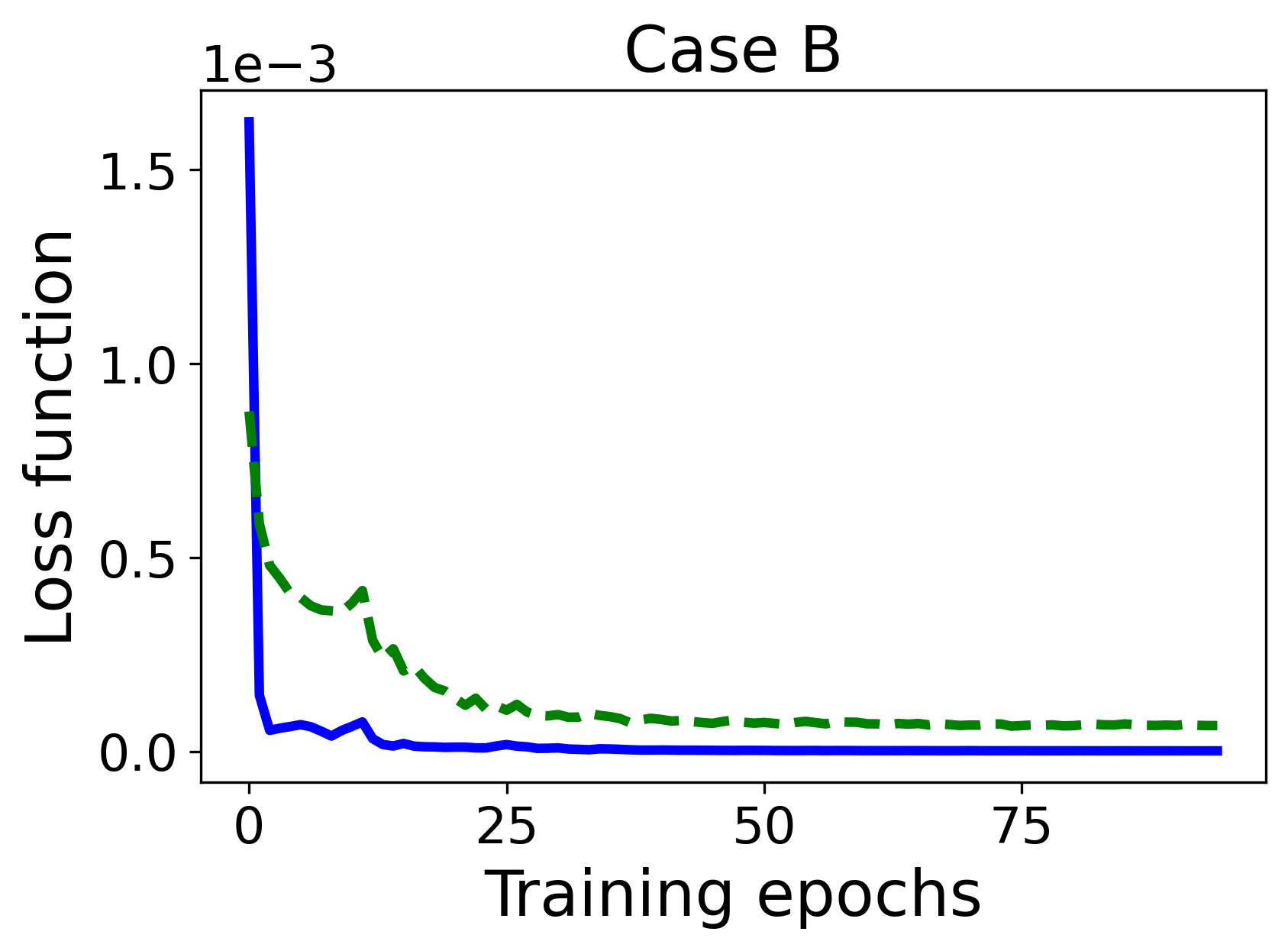}
	\includegraphics[height=0.3\textwidth]{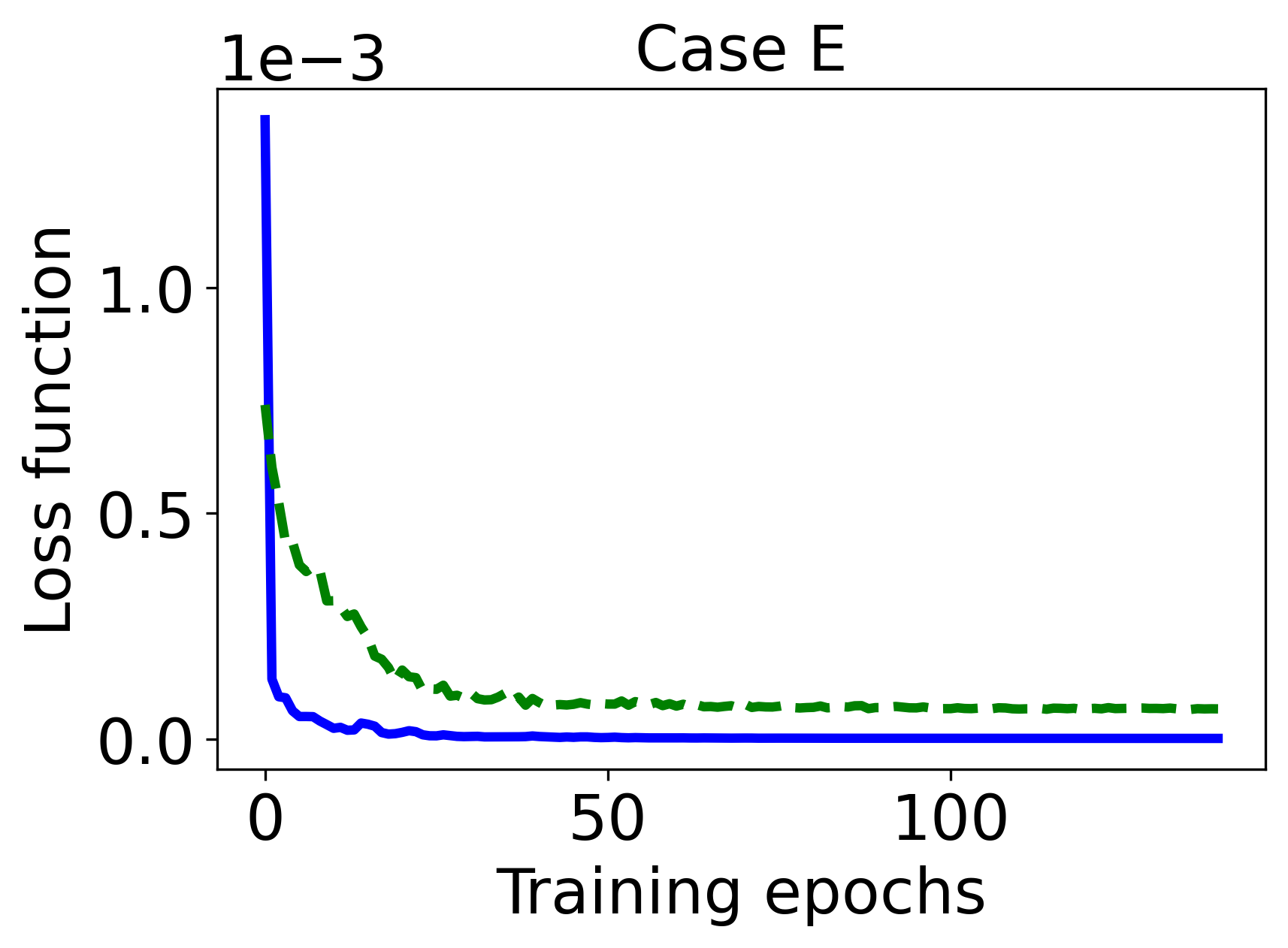}\\
	\caption{Evolution of the value of the loss function for the training set (solid blue) and validation set (dashed green) for the two best LSTM cases analyzed.}
	\label{fig:Figure3}
\end{figure}

\begin{figure}
	\centering
	\includegraphics[height=0.3\textwidth]{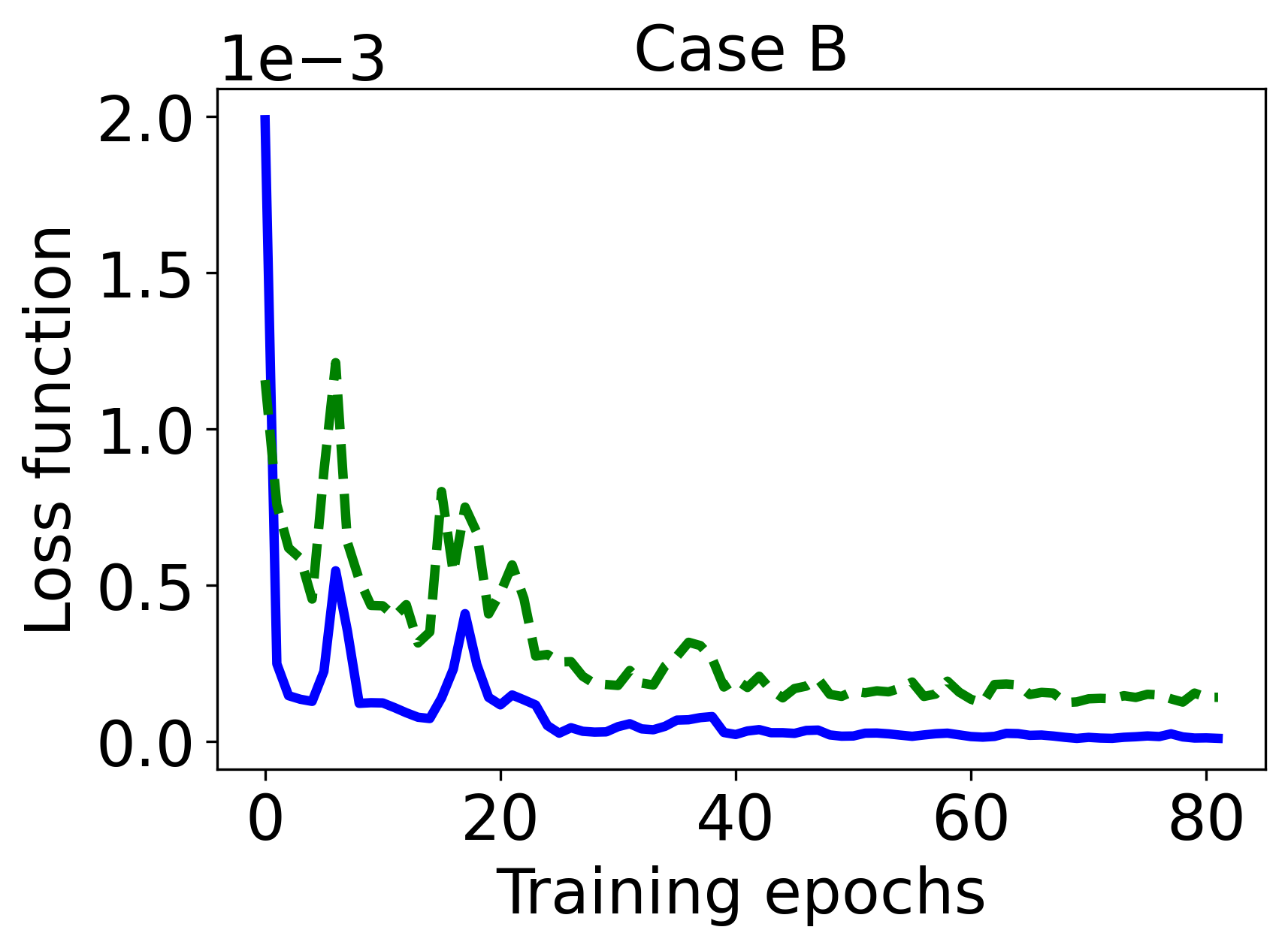}
	\includegraphics[height=0.3\textwidth]{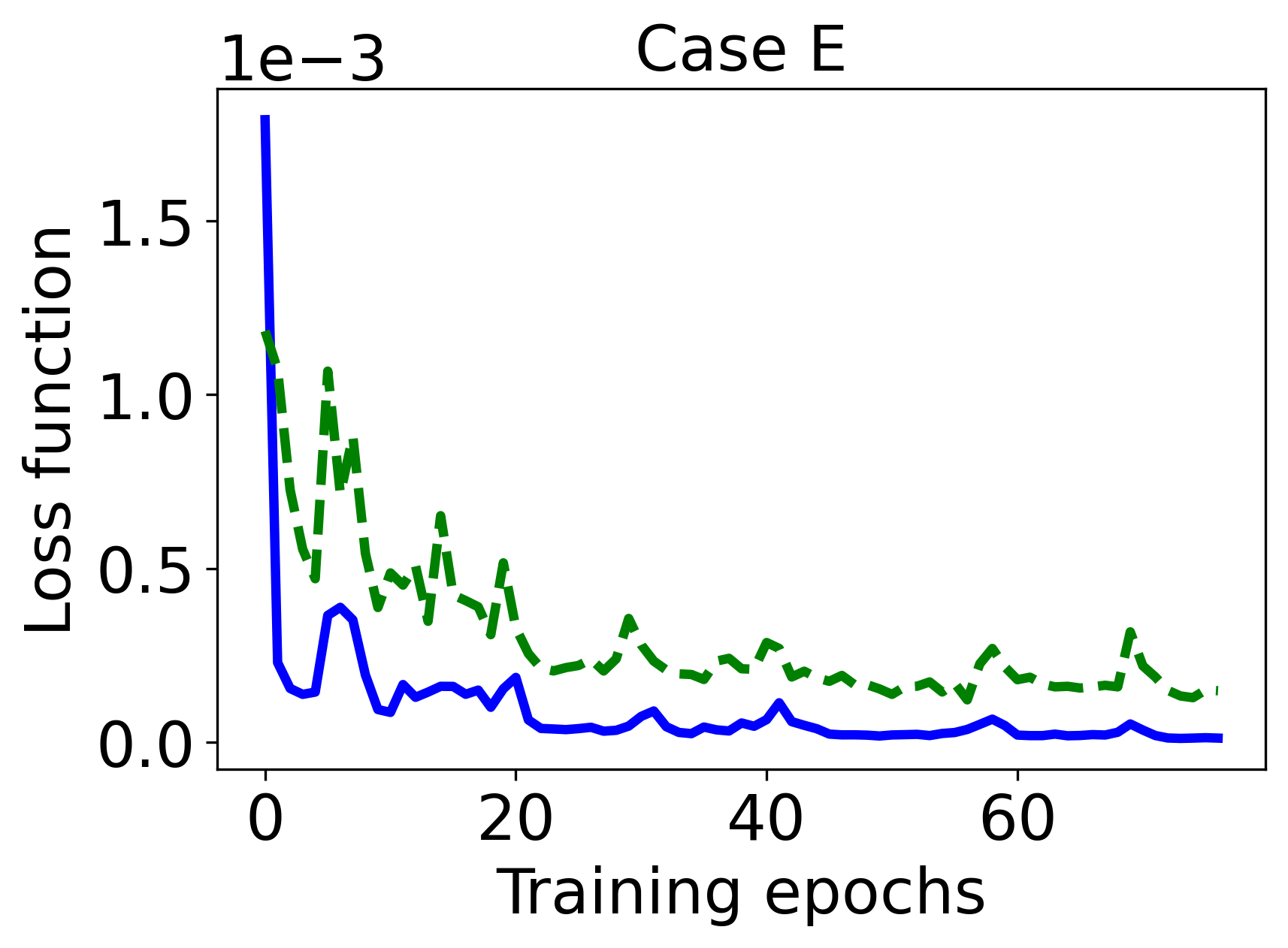}\\
	\caption{Evolution of the value of the loss function for the training set (solid blue) and validation set (dashed green) for the two best CNN cases analyzed.}
	\label{fig:Figure4}
\end{figure}

The training and validation sets have been used to train the two different deep learning models proposed in section \ref{sec:DLmodels}. The loss function decay for the best two cases for both LSTM and CNN (cases B and E) is plotted in Figs. \ref{fig:Figure3} and \ref{fig:Figure4}, respectively. The decay for the other cases is shown in \ref{App:A}, Figs. \ref{fig:FigureA1} and \ref{fig:FigureA2}. 
The value of the loss function of the original case (Case 0) decreases with the number of epochs, to values of the $MSE$ of order $10^{-3}$ in both LSTM and CNN models. 
Cases A and C don't seem to improve the performance of the original case, reaching values of $MSE$ of the same order.
Cases B, D, and E appear to be the best ones, since the loss function decreases with the number of epochs, reaching values of order $10^{-4}$ for the validation set. This is further confirmed by the RRMSE of the validation set, as shown in Tab. \ref{tab:RRMSE_val}. 

\begin{figure}
	\centering
	\includegraphics[height=0.28\textwidth]{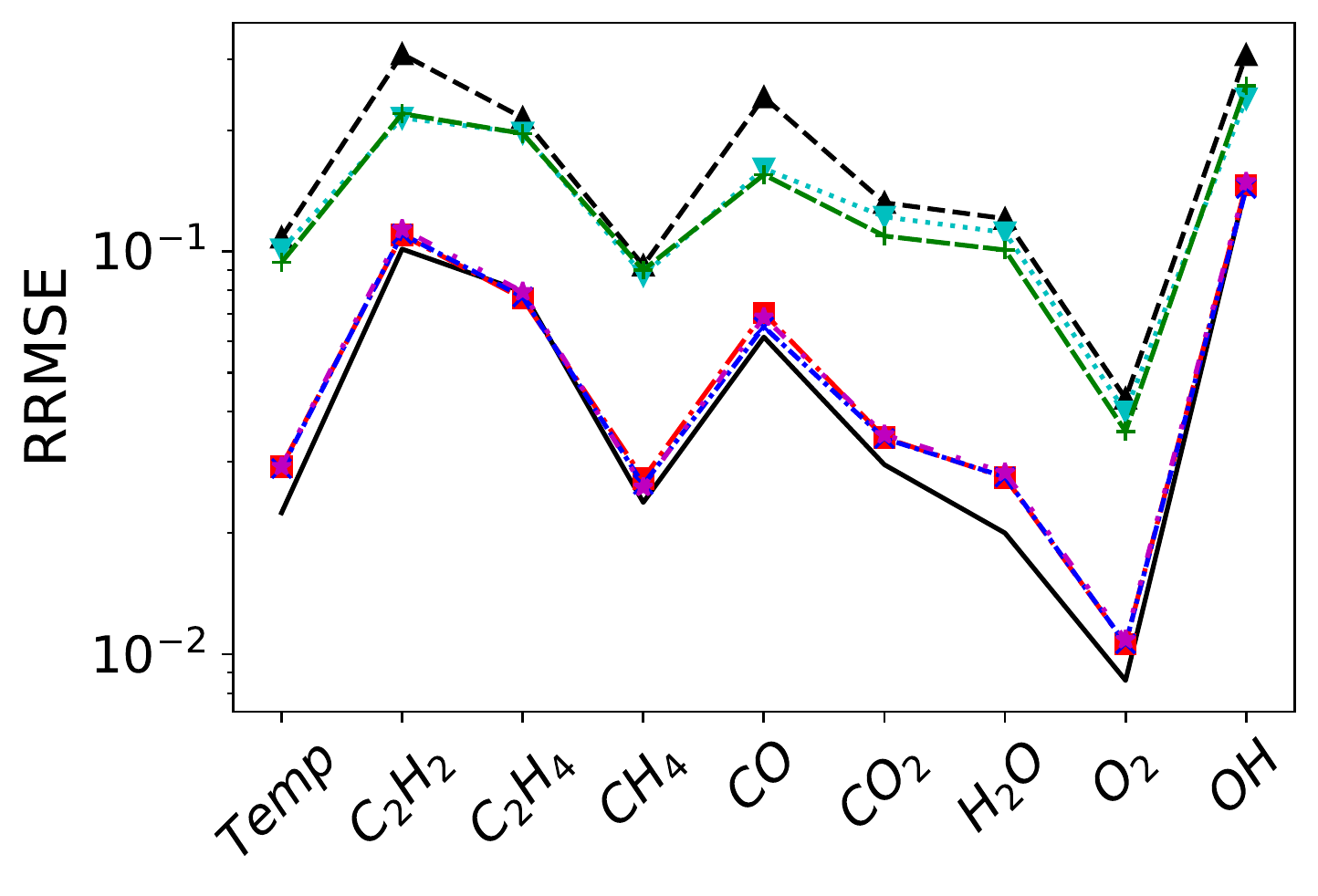}
	\includegraphics[height=0.28\textwidth]{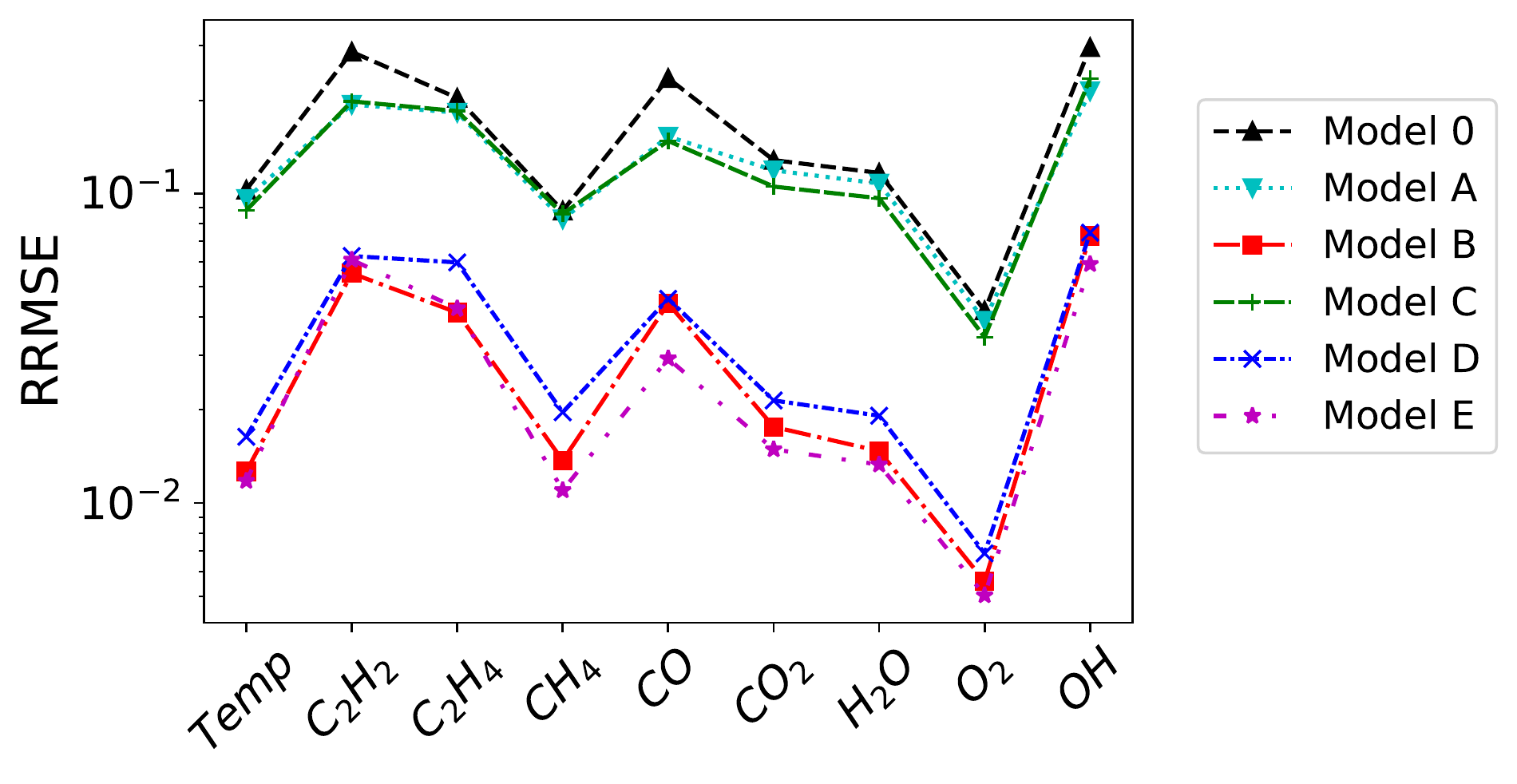}
	\caption{Prediction error for the variables studied using the different LSTM cases, compared with the original (left) and the reconstruction with SVD (right). The black line on the  left plot represents the reconstruction error using SVD.}
	\label{fig:Figure5}
\end{figure}

\begin{figure}
	\centering
	\includegraphics[height=0.28\textwidth]{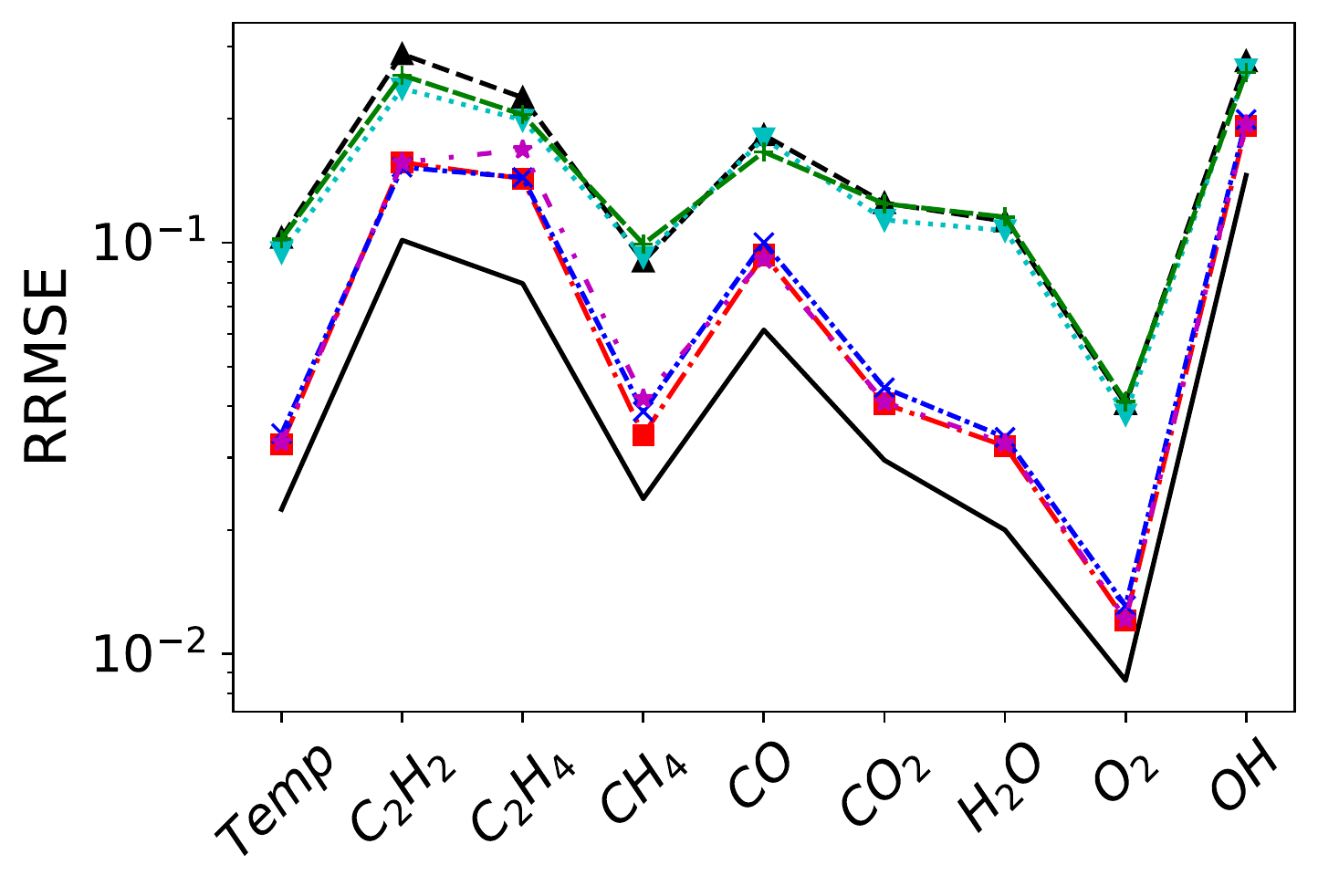}
	\includegraphics[height=0.28\textwidth]{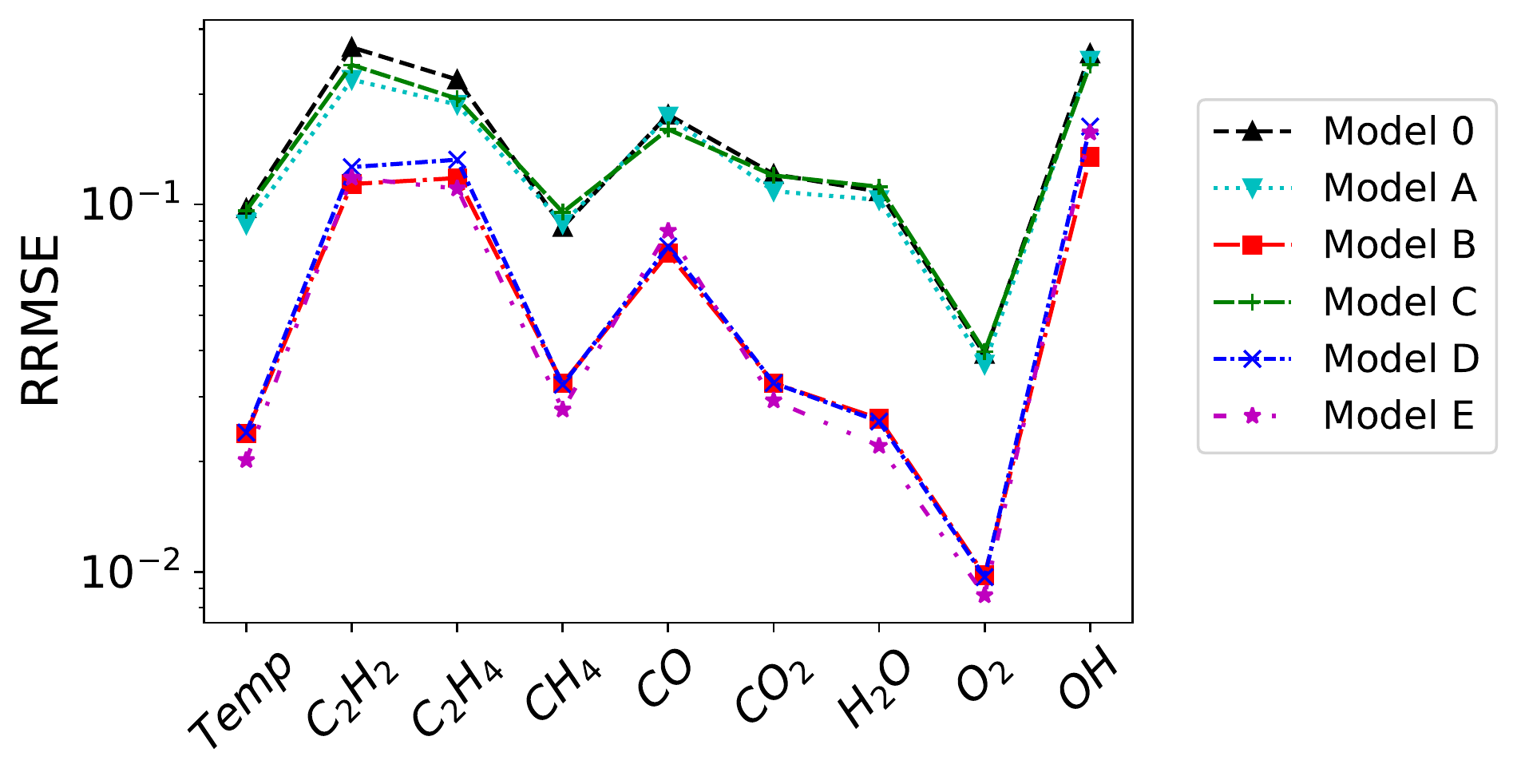}
	\caption{Prediction error for the variables studied using the different CNN cases, compared with the original (left) and the reconstruction with SVD (right). The black line on the left plot represents the reconstruction error using SVD.}
	\label{fig:Figure6}
\end{figure}

\begin{table}[H] 
	\centering
    \begin{tabular}{|c|c|c|c|c|c|c|}
    \hline
     & Case 0 & Case A & Case B & Case C & Case D & Case E  \\
     \hline
     LSTM & $0.111$ & $0.1116$ & $0.0326$ & $0.1021$ & $0.0337$ & $0.0313$ \\
     \hline
     CNN & $0.1062$ & $0.094$ & $0.0382$ & $0.1003$ & $0.0352$ & $0.0363$ \\
     \hline     
    \end{tabular}
    \caption{RRMSE of the validation set for all the studied cases.}
    \label{tab:RRMSE_val}
\end{table}
The prediction error (RRMSE) was compared to the reconstruction error with SVD for the different variables
Figures \ref{fig:Figure5} and \ref{fig:Figure6} show that similar results are obtained with both LSTM and CNN models. The original case shows the worst prediction error for all the variables.
Cases A and C do not improve the prediction error to a large extent. These two cases, along with the original one, are the cases without changes in the activation function of the hidden layers. The results show that in both cases overfitting occurred during the training process, as shown in Figs.~\ref{fig:FigureA1}~and~\ref{fig:FigureA2}.
Cases B, D, and E improve the performance of the original one, being their prediction errors close to the SVD reconstruction error for all the variables studied. All three cases change the activation function of the hidden layers, suggesting that the change of this parameter is important to obtain the best predictions. Comparing cases D and E, it can be seen that the modification of the learning rate improves the performance of the models. The change in the loss function from the default $MSE$ to the physics-aware loss function slightly improves the performance of the neural network. This can be seen by comparing the performance of cases B and E. The RRMSE is slightly improved for case E when compared to case B. The introduction of the physics-aware loss function might be crucial in an extrapolation scenario in which the mass balance must be conserved in each time instant. Therefore, it is maintained despite the slight improvement in the reconstruction error.


\begin{figure}
	\centering
	\includegraphics[height=0.28\textwidth]{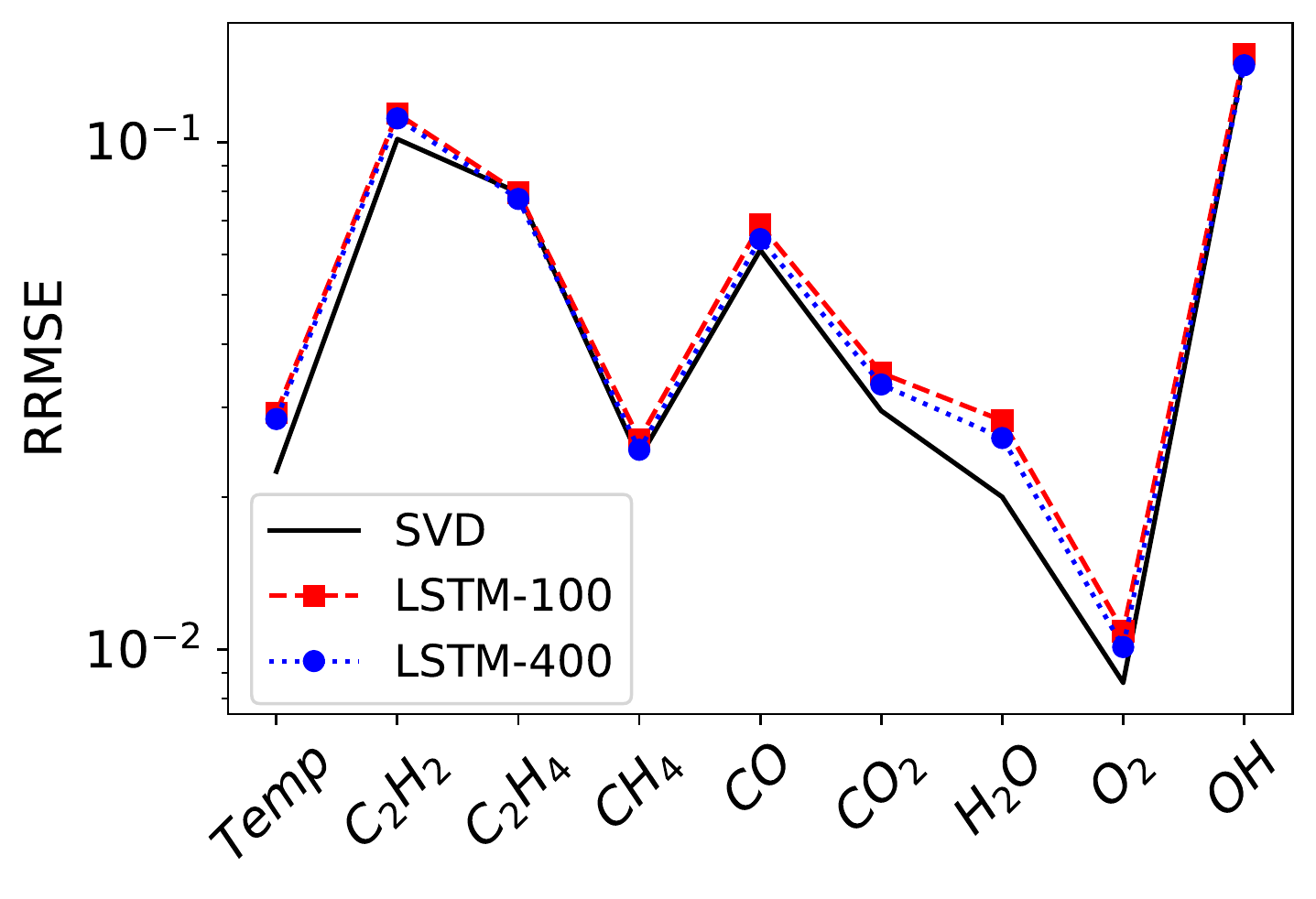}
	\includegraphics[height=0.28\textwidth]{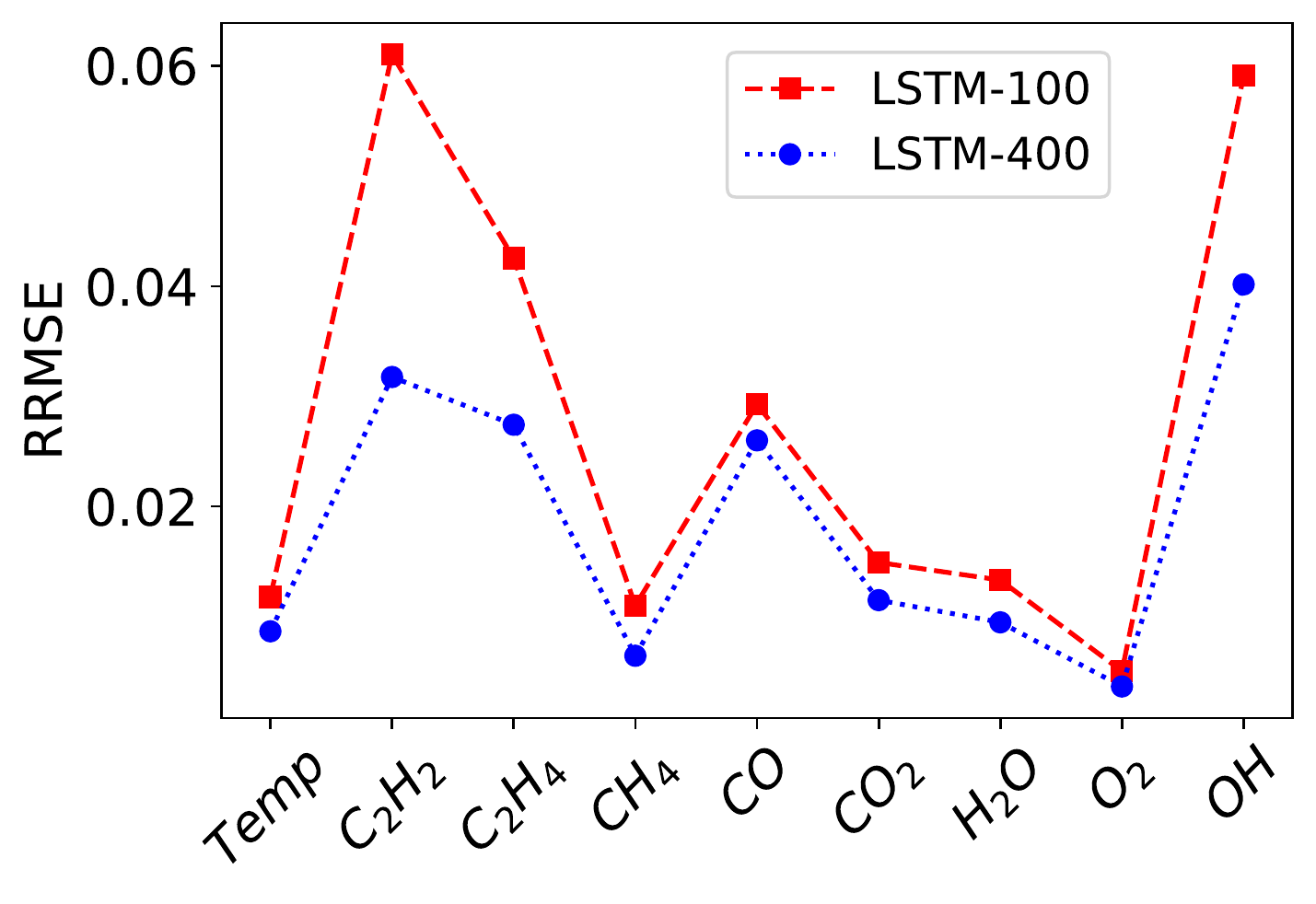}
	\caption{Prediction error for the variables studied using the LSTM architecture with the different numbers of neurons, compared with the original data (left) and the reconstruction with SVD (right). The black line on the left plot represents the reconstruction error using SVD.}
	\label{fig:Figure7}
\end{figure}

The number of neurons in the LSTM model has been analyzed. For that purpose, the number of neurons in the LSTM layer was increased from $100$ to $400$ on the best performing model, model E. The prediction error has been calculated for the different variables and compared to the reconstruction error using SVD. As seen in Fig. \ref{fig:Figure7}, the neural network with $400$ neurons slightly improves the prediction error, while the number of parameters on the architecture is increased from $125028$ to $1673628$ ($\sim 1238\%$), so the higher computational cost does not compensate the improve of the predictions.

\subsection{Comparison between models \label{sec:com}}

\begin{figure}
	\centering
	\includegraphics[height=0.28\textwidth]{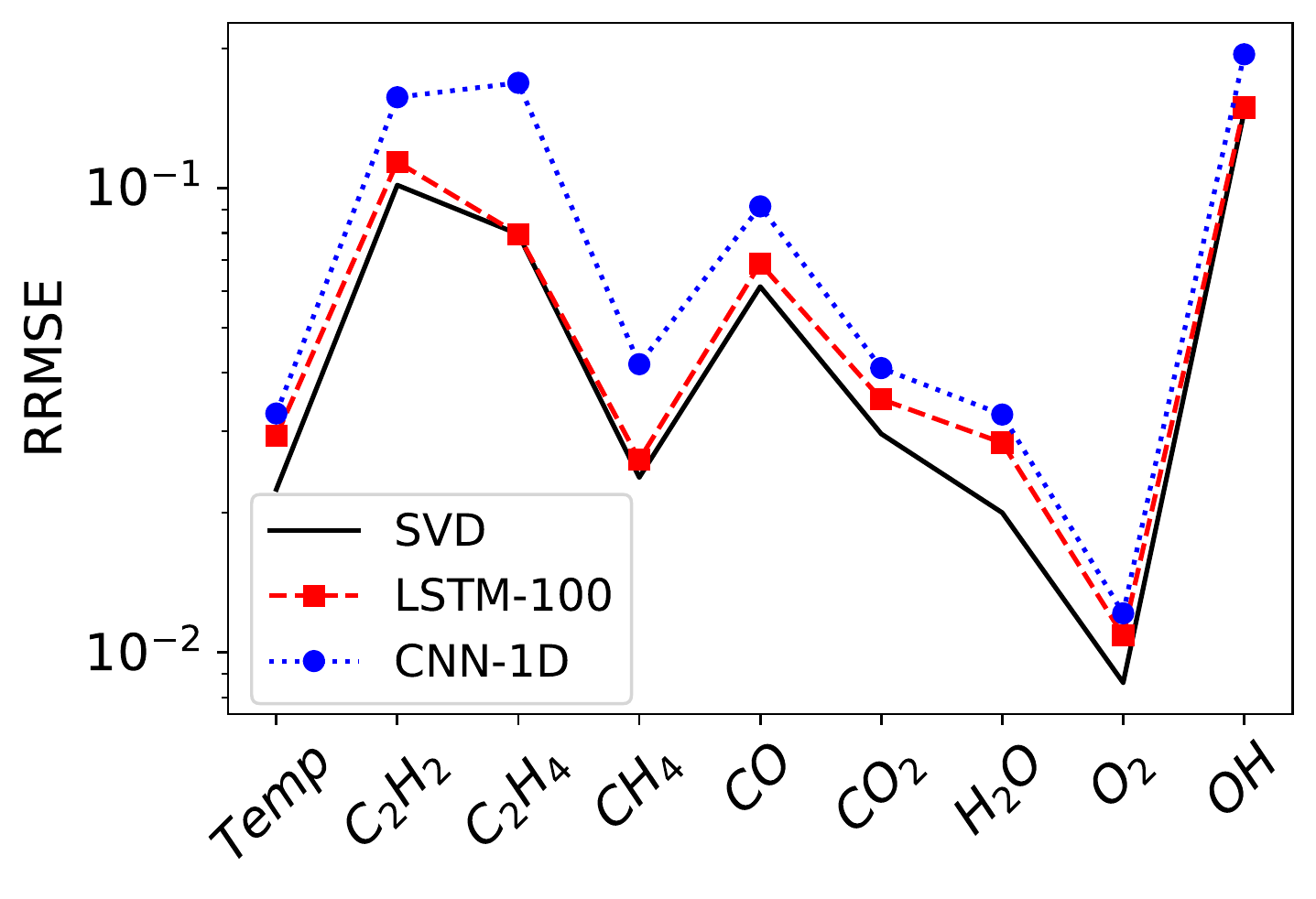}
	\includegraphics[height=0.28\textwidth]{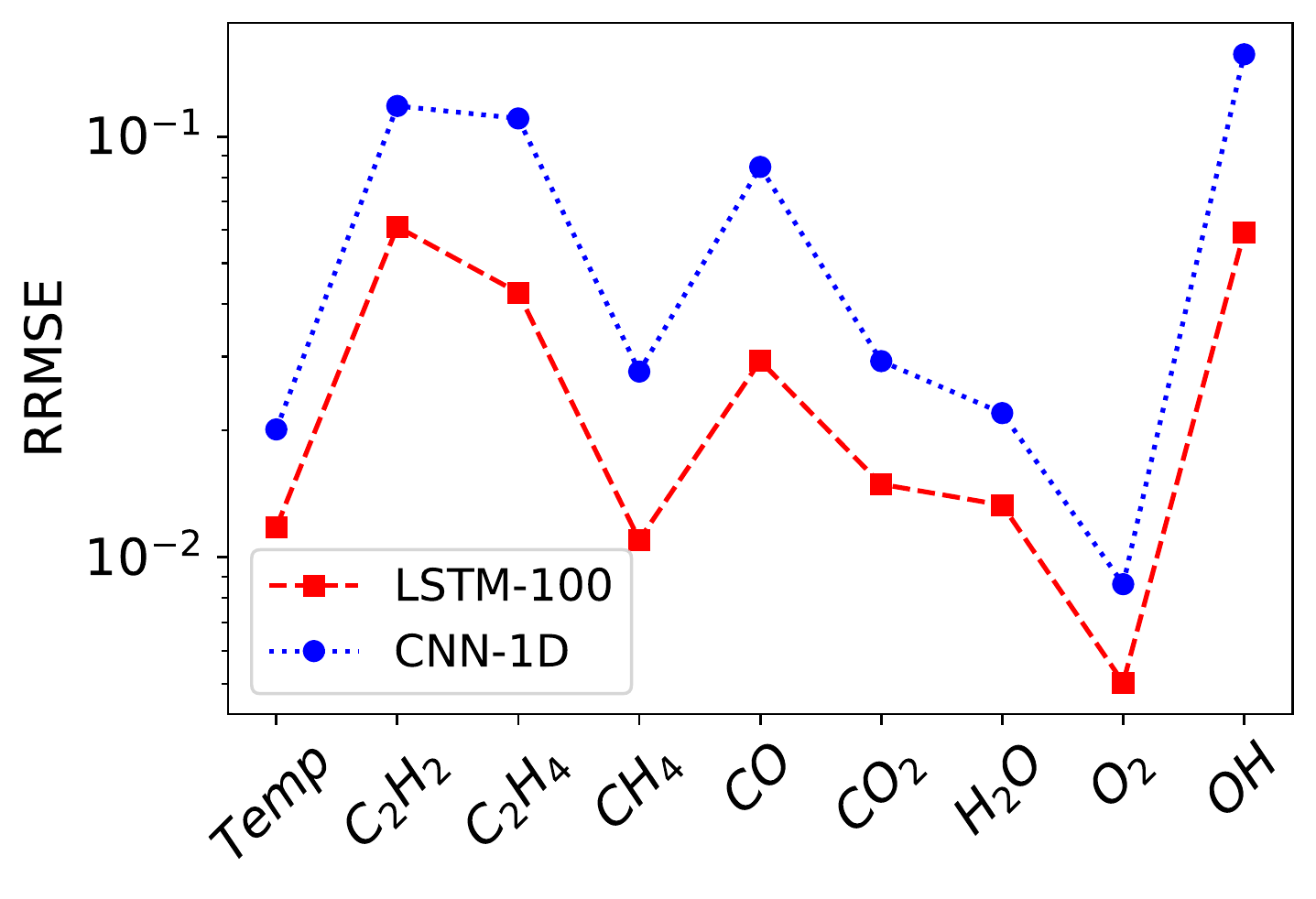}
	\caption{Prediction error for the variables studied using the different architectures (\textbf{dashed red:} LSTM neural network, \textbf{dotted blue:} CNN), compared with the original (left) and the reconstruction with SVD (right). The black line on the left plot represents the reconstruction error using SVD.}
	\label{fig:Figure8}
\end{figure}

Once the best hyperparameters have been selected, the LSTM and CNN models are compared. In Fig. \ref{fig:Figure8}, the prediction errors of the different variables have been plotted and compared with the reconstruction error using SVD. It is clear that the LSTM model provides the best predictions, being closer to the reconstruction error for all the variables, while the CNN shows higher error for all the variables analyzed. That might be partially explained by the feedback mechanism in the LSTM layer, making such a model more robust to reproduce temporal dynamics.

\begin{figure}
	\centering
	\includegraphics[width=0.95\textwidth]{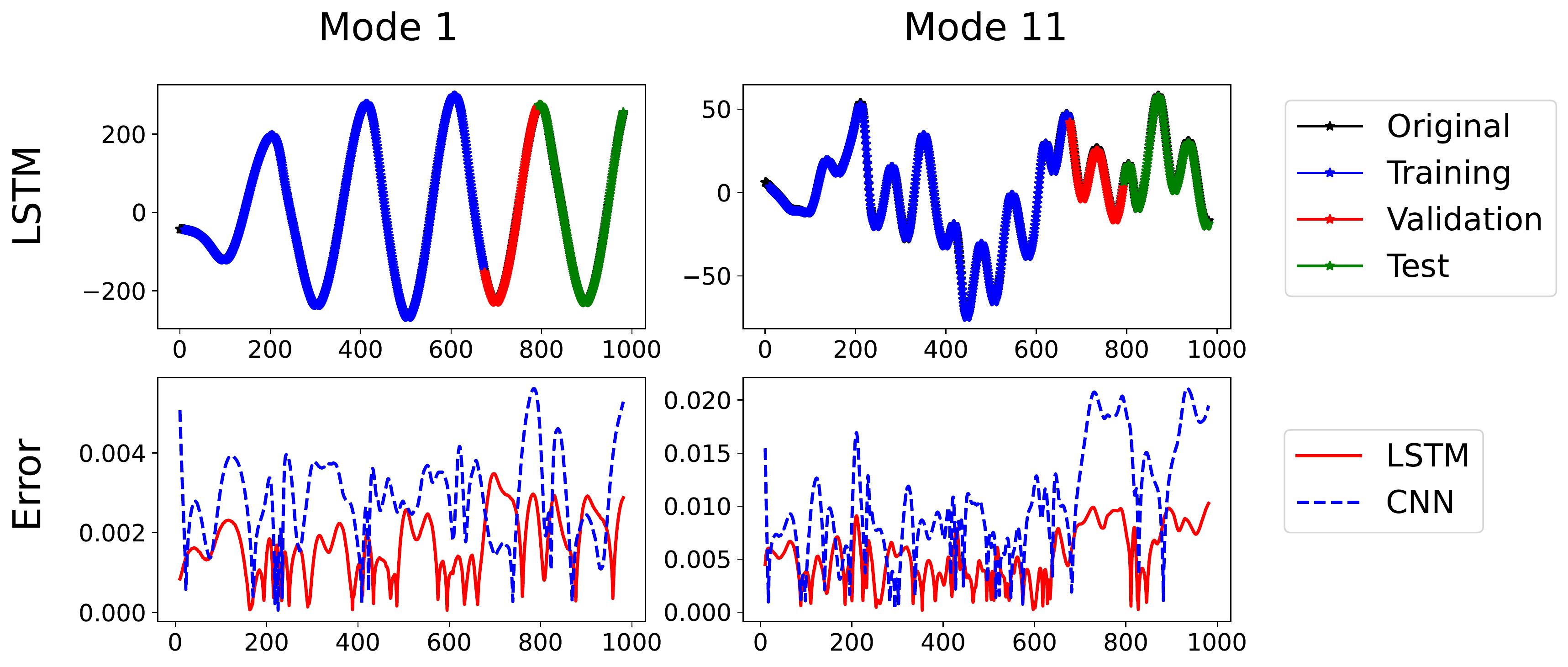}
	\caption{On top, prediction of the temporal modes $1$ and $11$ using the LSTM neural network. \textbf{Blue:} training set, \textbf{red:} validation set, \textbf{green:} test set and \textbf{black:} original mode. At the bottom, are the prediction errors of the mentioned modes, for both LSTM and CNN models. }
	\label{fig:Figure9}
\end{figure}

The predictions of the temporal modes have been analyzed in Figs. \ref{fig:Figure9} and \ref{fig:FigureA3}, against the original one. The selected modes to be studied are the first one and three randomly selected modes. For analyzing the predictions, the modes have been plotted, as well as the error has been calculated as

\begin{equation}
	n-MSE(t)=\frac{\|\hat{\bT}_{j,t}-\hat{\bT}_{j,t}^{*}\|}{\text{max}(\hat{\bT}_{j})-\text{min}(\hat{\bT}_{j})}, \label{eq:nMSE}
\end{equation}

\noindent where $\hat{\bT}_{j,t}$ is the original $j$-th mode at the $t$-th time instant and $\hat{\bT}_{j,t}^{*}$ is the prediction. As seen, the differences in the first mode, the one with periodical behavior, are negligible. Both models are able to correctly predict the mode. The error made during training is less than the one made at the validation and test, as expected. However, for these two sets, the error is maintained stable and below $0.5\%$. The error of the LSTM predictions is lower than the one from CNN. For the eleventh mode, the error is larger than for the first one. Nevertheless, the LSTM model is able to predict it with an error below $1\%$. Other modes are predicted with less accuracy, such as mode $18$ (Figure \ref{fig:FigureA3}). The error is below $8\%$ for the CNN model and below $4\%$ for the LSTM model.

\begin{figure}
	\centering
	\includegraphics[width=0.65\textwidth]{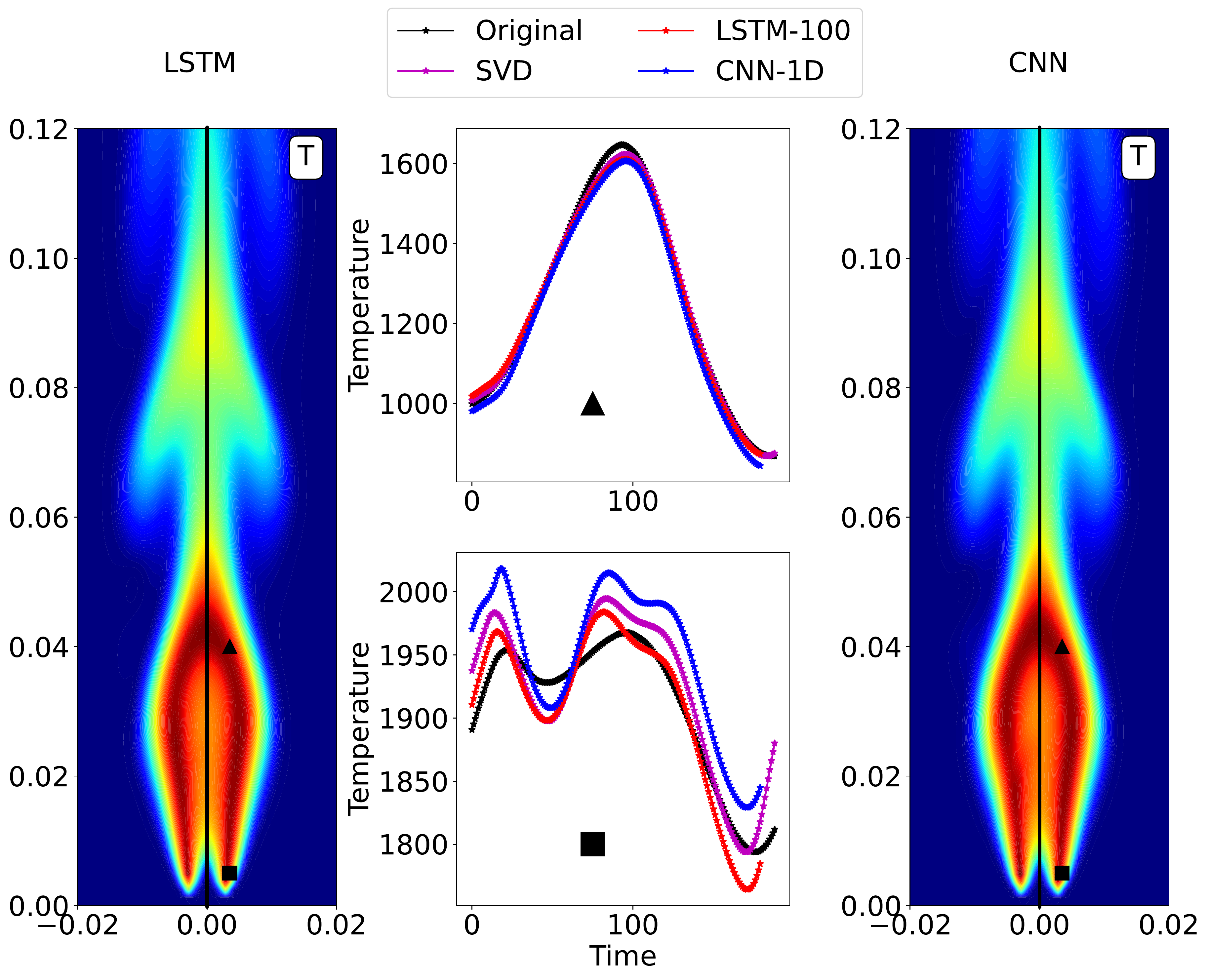}
	\caption{Contours of the prediction of a representative snapshot of the temperature with the LSTM (left) and CNN (right) models. The right part of each contour is the original snapshot and on the left is the prediction. In the middle part of the figure: Evolution of the prediction of the temperature at two characteristic points. Points extracted where the triangle and square are located in the figure contour.}
	\label{fig:Figure10}
\end{figure}

\begin{figure}
	\centering
	\includegraphics[width=0.65\textwidth]{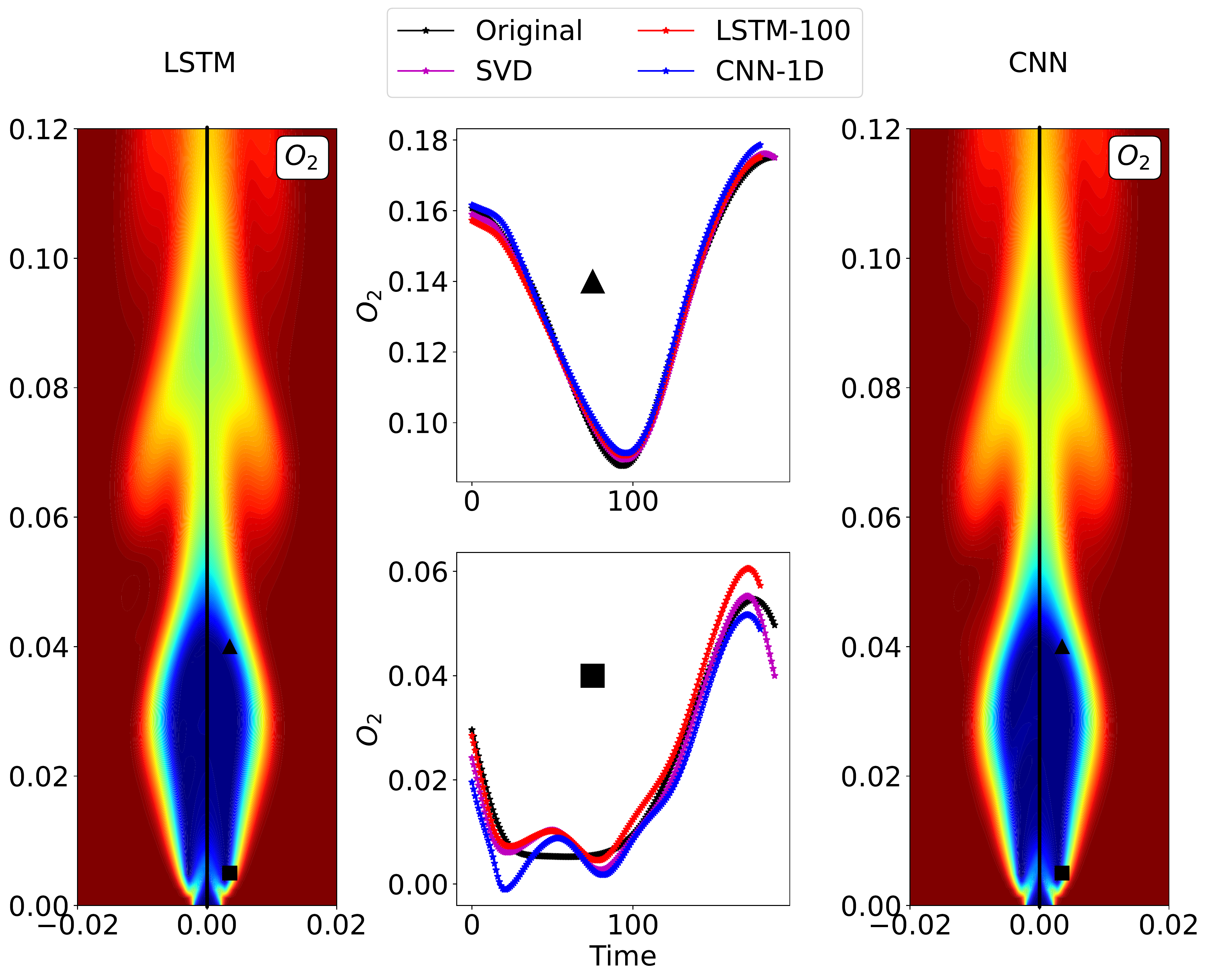}
	\caption{Same as Fig. \ref{fig:Figure10} for the $O_2$ mass fraction.}
	\label{fig:Figure11}
\end{figure}

\begin{figure}
	\centering
	\includegraphics[width=0.65\textwidth]{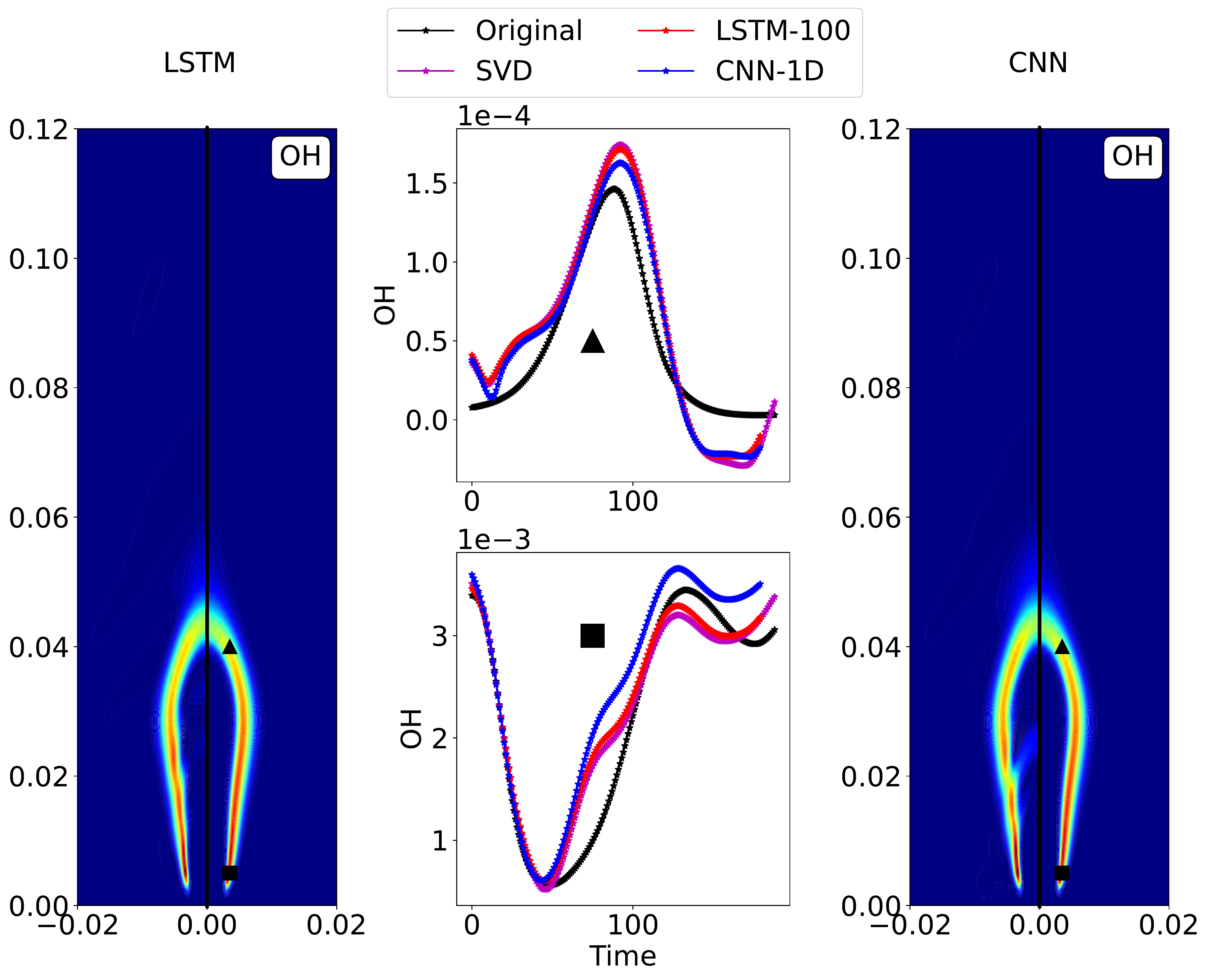}
	\caption{Same as Fig. \ref{fig:Figure10} for the $OH$ mass fraction.}
	\label{fig:Figure12}
\end{figure}

Figures \ref{fig:Figure10}, \ref{fig:Figure11} and \ref{fig:Figure12},  show a representative snapshot of the temperature, $O_2$ and $OH$ mass fractions are compared from a qualitative perspective, as well as the evolution in time of two characteristic points are plotted.

The temperature is satisfactorily predicted by both deep learning models, as shown in Fig. \ref{fig:Figure10}. There are no significant differences in the contours. Analyzing the evolution of the two characteristic points in time, it is seen that the differences on the far field are insignificant. Near the exit of the nozzle, slightly larger discrepancies are visible. The SVD algorithm reconstructs the original data fairly well. Furthermore, the prediction with the LSTM model is more accurate than the one with the CNN model. 
Similar conclusions can be made from the analysis of the $O_2$ mass fraction (Fig. \ref{fig:Figure11}).

The differences are more noticeable for the $OH$ mass fraction, particularly for the CNN model. While the main shape of the chemical species is well preserved, near the symmetry axis, the $OH$ profile is strongly underestimated. The evolution in time shows that the predictions by the deep learning models are close to the reconstruction with SVD, particularly using LSTM, but larger errors are observed between the SVD and the original data. 

\subsection{Transfer learning}

This section intends to explore the ability of the best previously trained deep learning model to predict a laminar flame under new conditions. The new database comes from the same configuration as in Sec. \ref{sec:Sim}, although the fuel is injected with a more complex velocity profile, given as

\begin{equation}
	v(r,t) = v_{max} \left( 1 - \frac{r^2}{R^2} \right) [1 
	+ \left( A_1 \ sin(2 \pi f_1 \ t)
	+  A_2 \ sin(2 \pi f_2 \ t)
	+  A_3 \ sin(2 \pi f_3 \ t)\right)/3], \label{eq:complexProfile}
\end{equation}

\noindent where $v_{max} = 70$ cm/s is the maximum velocity, $r$ is the radial coordinate and $R$ the internal radius of the nozzle. Equation \ref{eq:complexProfile} indicates that now there are three perturbations present in the profile, with frequencies $f_1 = 10 Hz$, $f_2 = 40 Hz$ and $f_3 = 80 Hz$, and amplitudes $A_1 = 0.9$, $A_2 = 0.5$ and $A_3 = 0.75$. More information about the numerical setting and the kinetic mechanism can be found in Ref. \cite{d2020adaptive,d2020impact}. 

The case is more complex as the boundary condition of the fuel perturbation in time $t$ is a linear combination of harmonic functions with different amplitudes and frequencies, while only one frequency was present in the original case. From the simulation, the temperature and 9 chemical species ($O$, $O_2$, $OH$, $H_2O$, $CH_4$, $CO$,  $CO_2$, $C_2H_2$ and $N_2$) have been extracted each $\Delta t = 2.5 \times 10^{-4}$ (a total of $1000$ snapshots) and in a structured mesh of dimensions $300 \times 75$.

\begin{figure}
	\centering
	\includegraphics[width=0.95\textwidth]{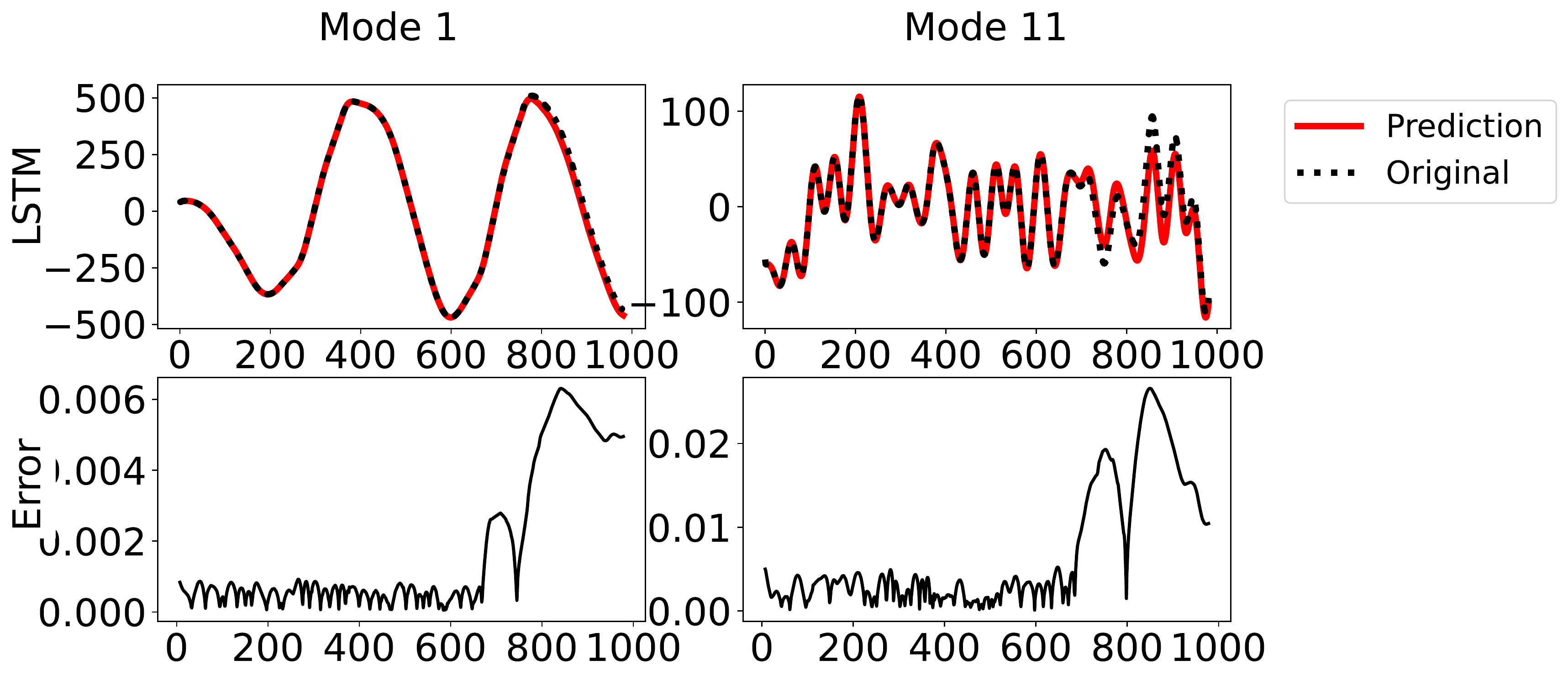}
	\caption{On top, prediction of the temporal modes $1$ and $11$ using the LSTM neural network. The prediction errors of the mentioned modes are plotted at the bottom part.}
	\label{fig:Figure14}
\end{figure}

Firstly, SVD is applied, and the first $18$ modes are selected, which account for the $81.3\%$ of the total energy. The LSTM model is used since it is the one that returns the best results. As the neural network has already been trained, all the dataset is labeled as a test. Figure \ref{fig:Figure14} shows the predictions of the first and eleventh modes, as well as the prediction errors, given by Eq.~\eqref{eq:nMSE}. The first mode is predicted with an error below the $0.64\%$. The eleventh mode is also correctly predicted with a maximum error below $3\%$.

\begin{figure}
	\centering
	\includegraphics[width=0.65\textwidth]{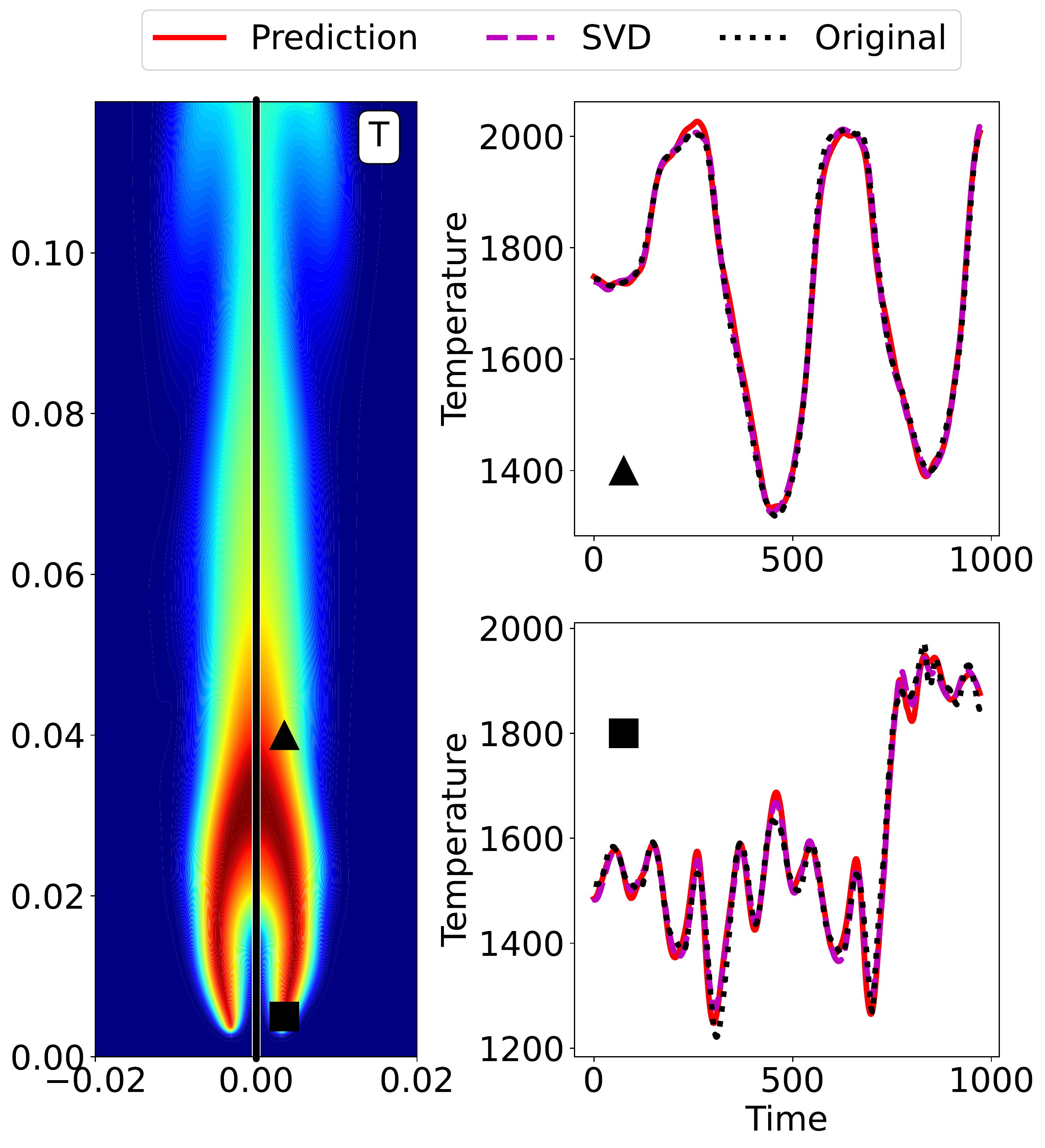}
	\caption{\textbf{Left:} Contour of the prediction of a representative snapshot of the temperature. The right part of the contour is the original snapshot and on the left is the prediction. \textbf{Right:} Evolution of the prediction of the temperature at two characteristic points. Points extracted where the triangle and square are located in the figure contour.}
	\label{fig:Figure15}
\end{figure}

Figure~\ref{fig:Figure15} illustrates a representative snapshot of the temperature and the evolution in time of two characteristic points is plotted. From a qualitative perspective, few differences are noticeable when comparing the original and predicted snapshots. The evolution in time of the point far from the injector nozzle, where the dynamic has a smoother behavior, is predicted with high accuracy. There are no qualitative differences between the SVD reconstruction, the original evolution in time, and the prediction using the deep learning model. Analyzing now the prediction of the point near the injector nozzle, the square in Fig.\ref{fig:Figure15}, some differences between the original evolution and the reconstruction with SVD can be noticed. The prediction error remains below $2\%$, although the dynamic has more complex behavior. From these results, it can be concluded that the trained neural network is valid for the prediction of more complex laminar flames. Furthermore, these results suggest that such a model might be a viable route to construct ROMs for turbulent reacting flows.

\section{Conclusions \label{sec:conclusions}}

This article presents a novel physics-aware predictive ROM for reacting flows. This ROM combines POD for dimensionality reduction with neural networks to predict the temporal coefficients. 

The algorithm consists of three main steps. First, each variable is centered and scaled with the standard deviation (\textit{auto scaling}). This step has to be applied due to the multivariate nature of reacting flows. Then, SVD is applied for obtaining the POD modes. The largest scales are retained for dimensionality reduction. Lastly, the temporal coefficients are introduced into deep learning architectures to predict the time evolution of the coefficients. Multiplying them with the POD modes allows the prediction of the snapshots. Different cases based on LSTM and one-dimensional convolutional neural networks have been analyzed to study the influence of several parameters. The presented results can be summarized in the following points:

\begin{itemize}
    \item The models have been successfully applied and the activation functions are the most important parameters to improve the predictions in the neural network.
    Scaling improves the performance of the neural network on nonperiodic temporal modes. 
    
    \item The change in the activation functions makes the greatest improvement in the reconstruction error, close to the one with SVD. This reconstruction error is important as the algorithm predicts the temporal coefficients of a number of selected POD modes, which contain the largest scales, although not all scales, so some differences have to be made.
    
    \item The analysis of the prediction error allows to notice that the LSTM neural network returns better predictions than the convolutional one particularly for minor species.
    
    \item The LSTM model was found to be suitable for the prediction of different boundary conditions of the laminar flame. Only by training the neural network once, the model can predict the evolution in time of laminar flames with more complex velocity perturbations on the inlet. This last result shows the good capabilities of the model presented for transfer learning.

\end{itemize}

We place our contribution in the emerging area of physics-aware machine learning, where the final model, in many different ways blends two main components: availability of experimental data and/or often expensive computational models, and deep learning data-driven techniques. Such a combination allows understanding the flow physics of reacting flows at a relatively low cost, and also offers a broad spectrum of opportunities to leverage CFD codes.

The proposed algorithm can be employed in future works along with different non-linear modal decomposition, with the use of autoencoders to reduce the dimensionality of the data. Moreover, different input conditions of the same flame can be studied as part of future works, allowing the neural network to be suitable for time predictions of all the different conditions, with the training of just one case.

\section*{Acknowledgements}
A.C. and S.L.C. acknowledge the grant PID2020-114173RB-I00 funded by MCIN/AEI/ 10.13039/501100011033. S.L.C. and A.C. acknowledge the support of Comunidad de Madrid through the call Research Grants for Young Investigators from Universidad Politécnica de Madrid. A.C. also acknowledges the support of Universidad Politécnica de Madrid, under the program ‘Programa Propio’. Also, this work has received funding from the European Union’s Horizon 2020 research and innovation program under Marie Skłodowska-Curie grant agreement No $801505$.

\bibliography{sample}

\appendix
\section{}\label{App:A}

\begin{figure}[H]
	\centering
	\includegraphics[height=0.3\textwidth]{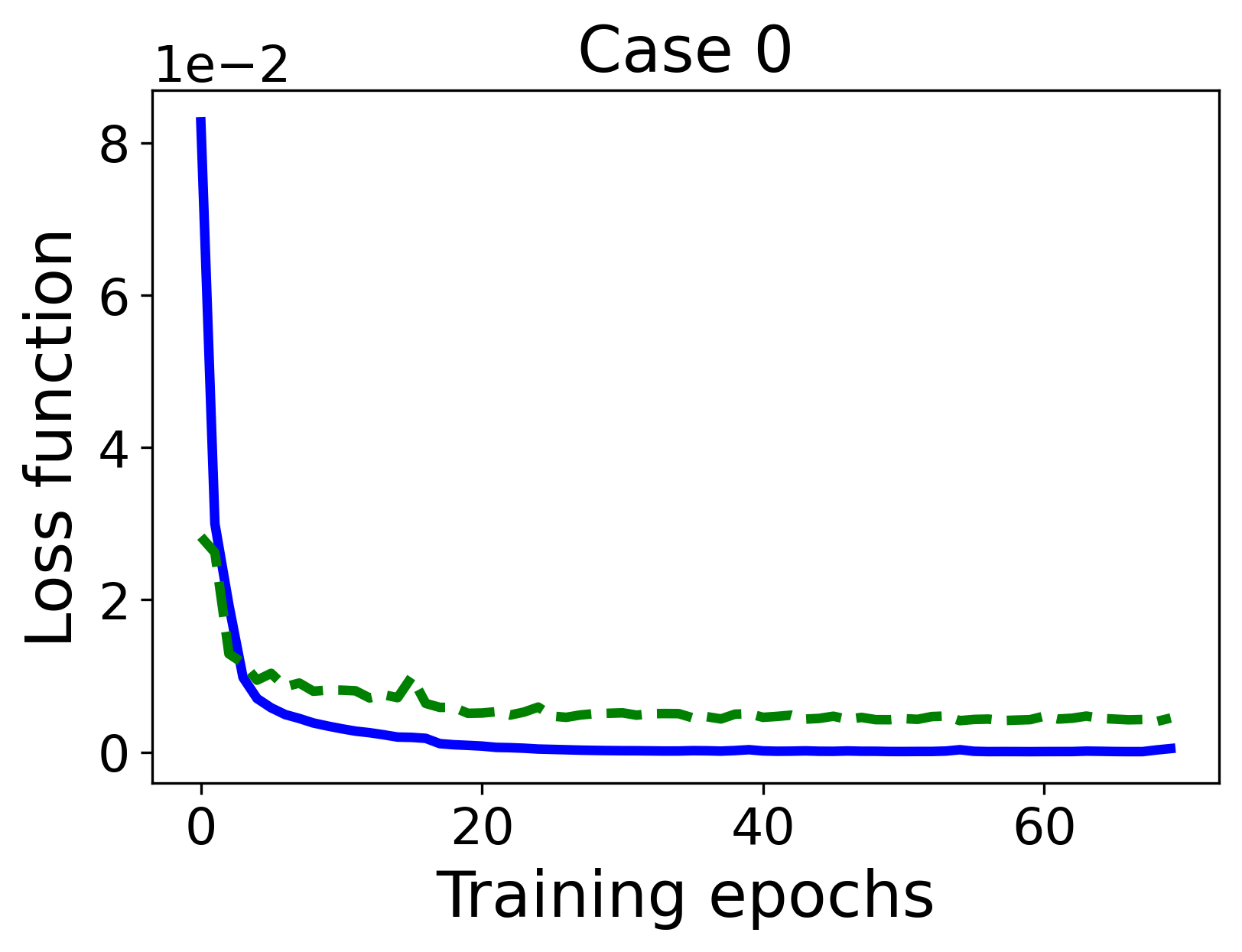}
	\includegraphics[height=0.3\textwidth]{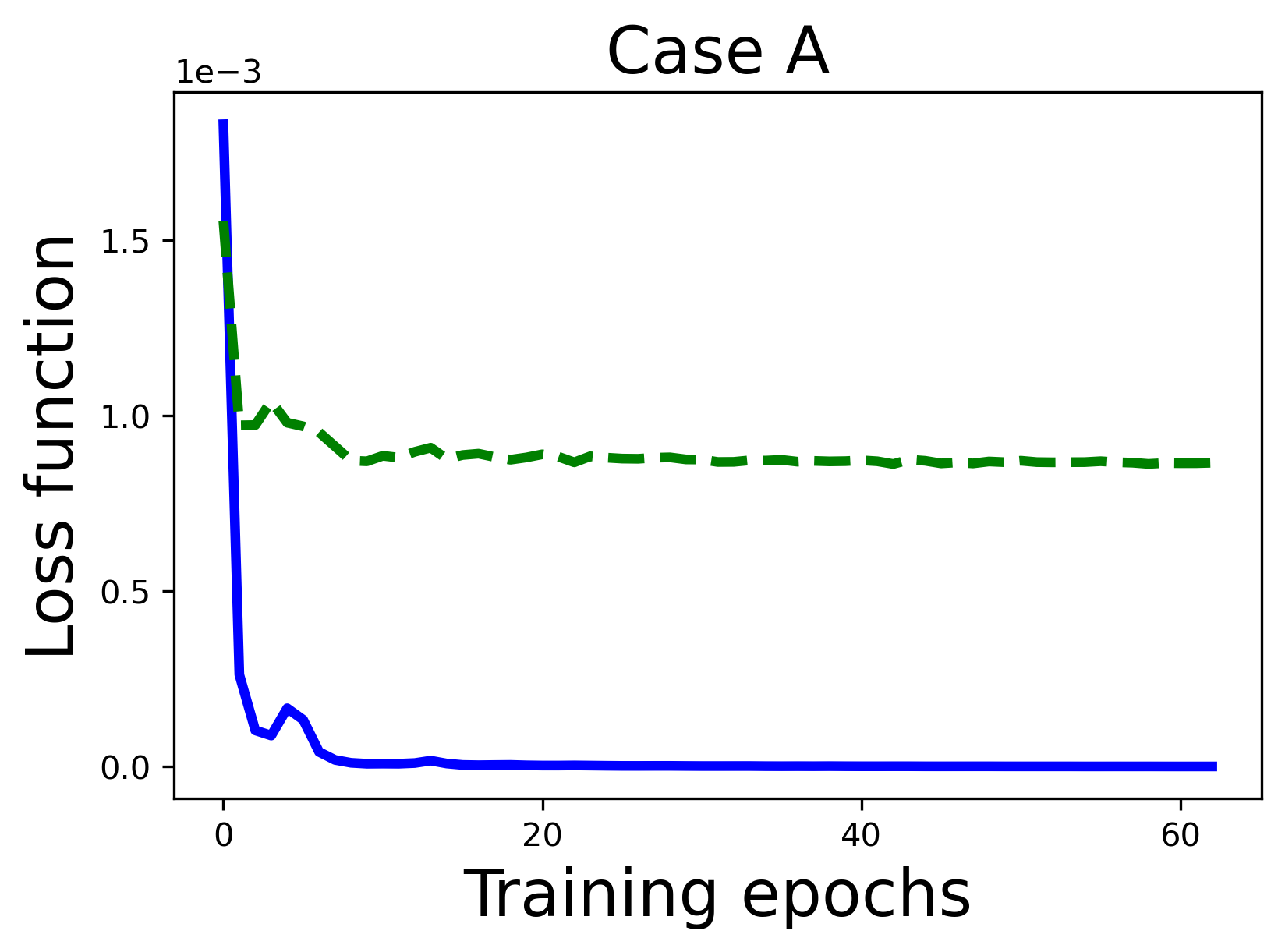}\\
	\includegraphics[height=0.3\textwidth]{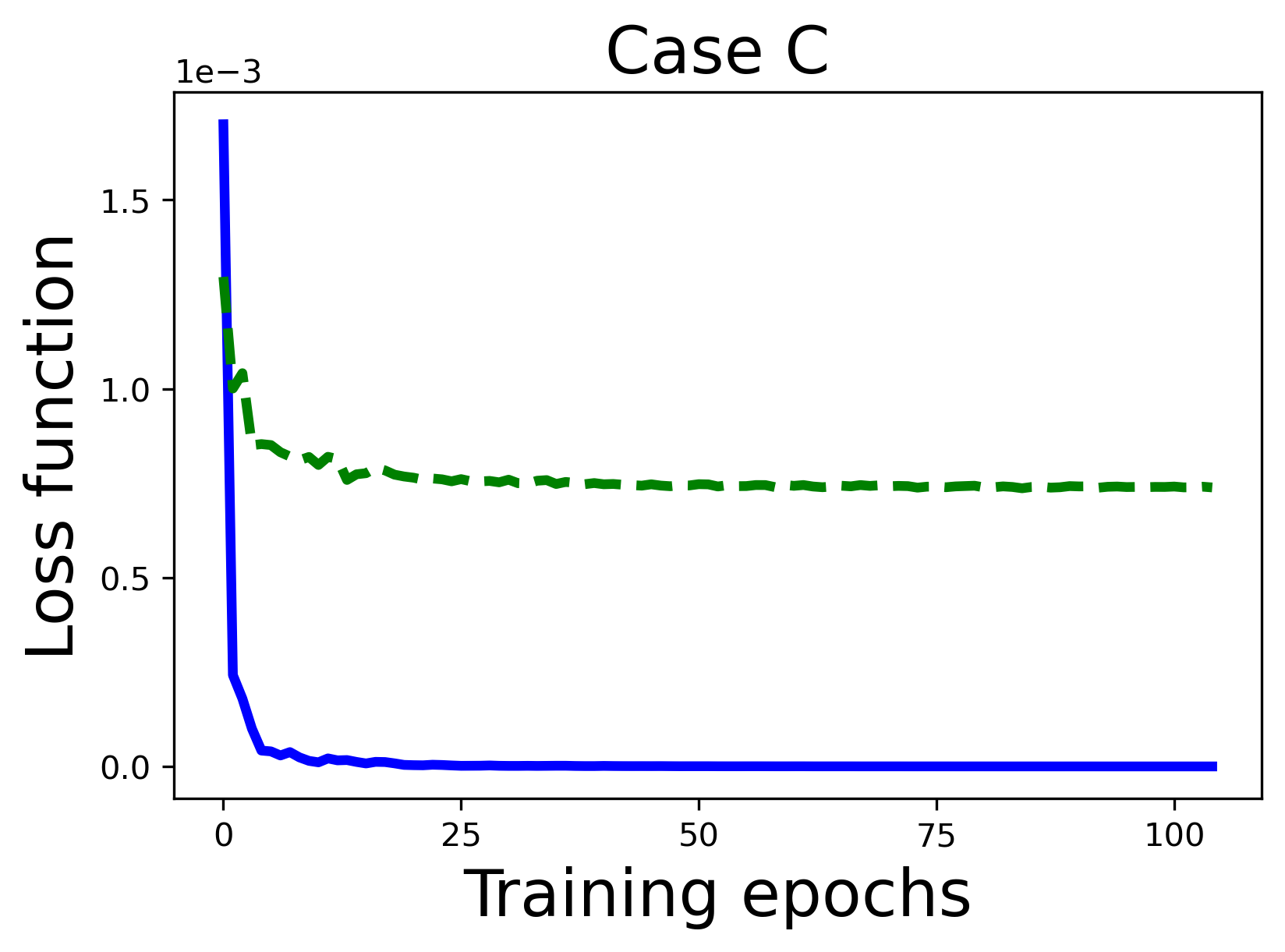}
	\includegraphics[height=0.3\textwidth]{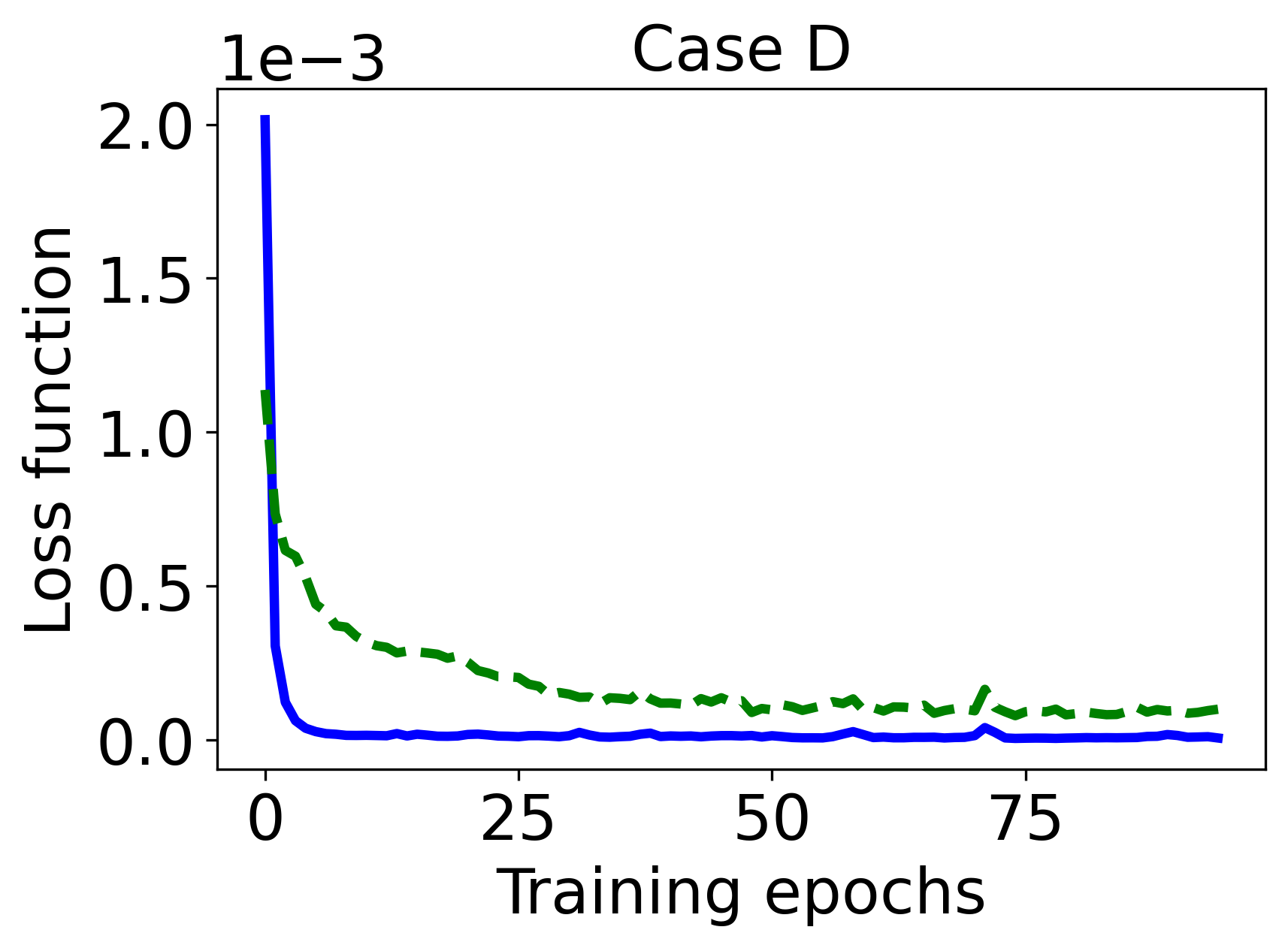}
	\caption{Evolution of the value of the loss function for the training set (solid blue) and validation set (dashed green) for the rest of the LSTM cases analyzed.}
	\label{fig:FigureA1}
\end{figure}

\begin{figure}[H]
	\centering
	\includegraphics[height=0.3\textwidth]{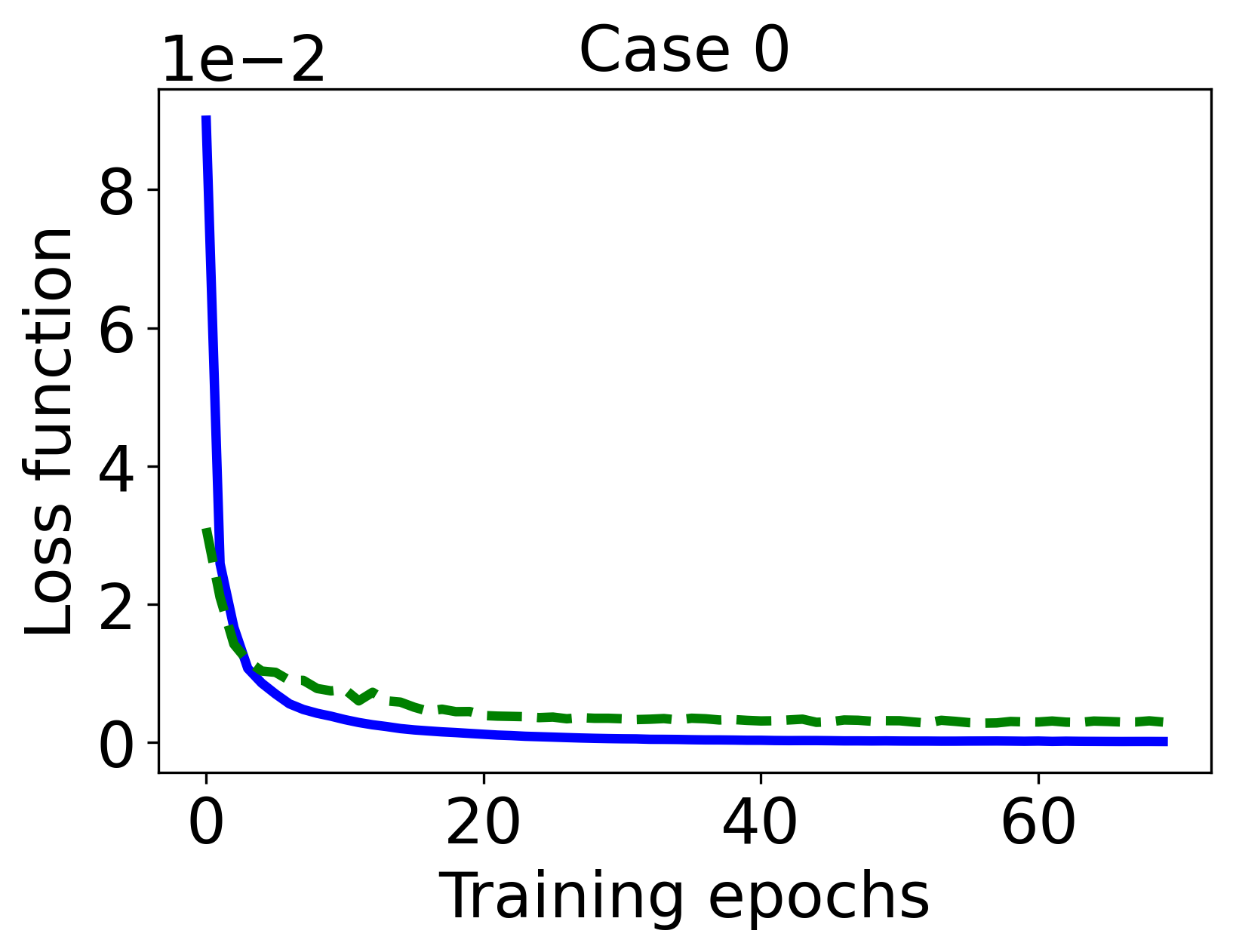}
	\includegraphics[height=0.3\textwidth]{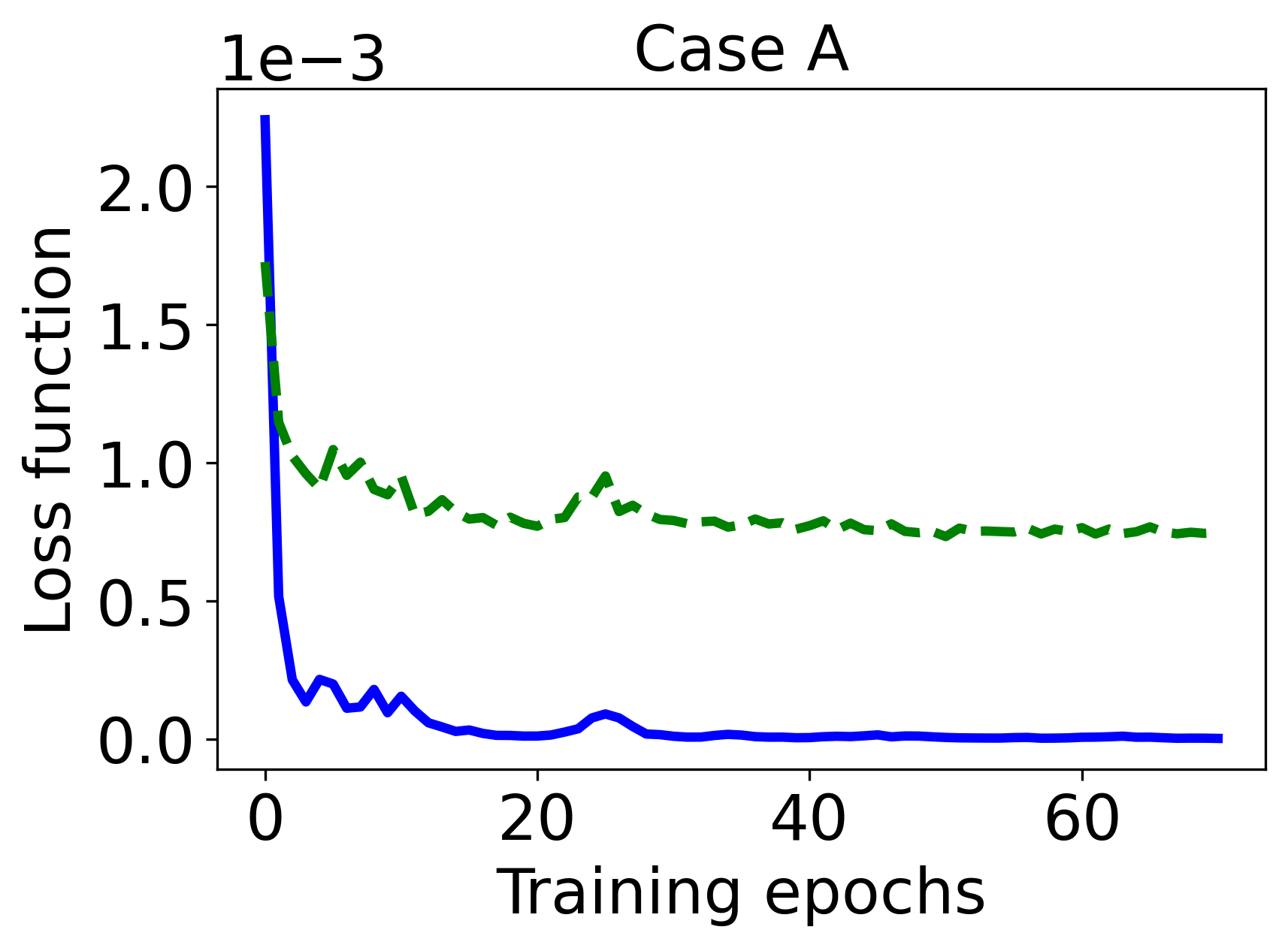}\\
	\includegraphics[height=0.3\textwidth]{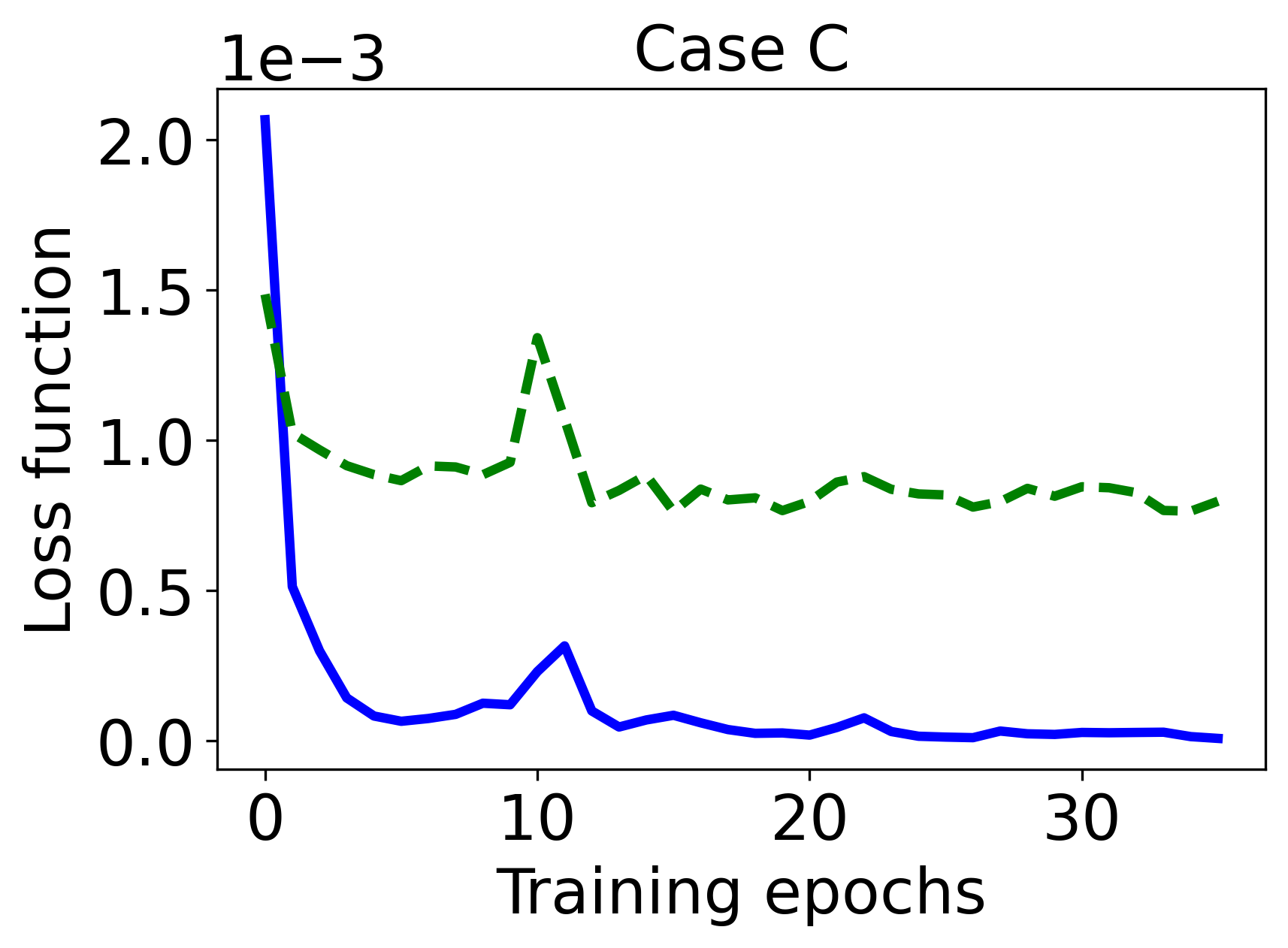}
	\includegraphics[height=0.3\textwidth]{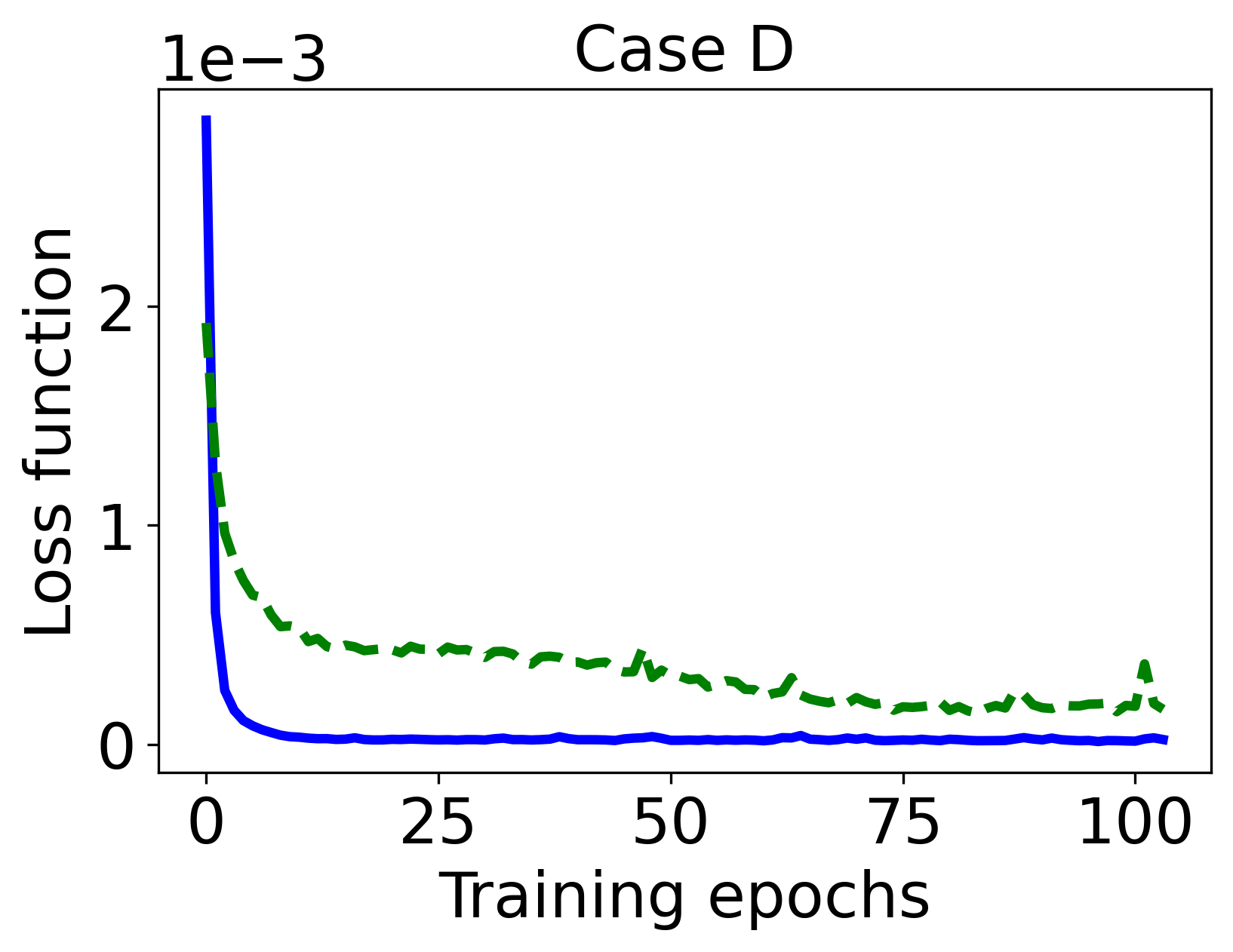}
	\caption{Evolution of the value of the loss function for the training set (solid blue) and validation set (dashed green) the rest of the CNN cases analyzed.}
	\label{fig:FigureA2}
\end{figure}
\begin{figure}[H]
	\centering
	\includegraphics[width=0.95\textwidth]{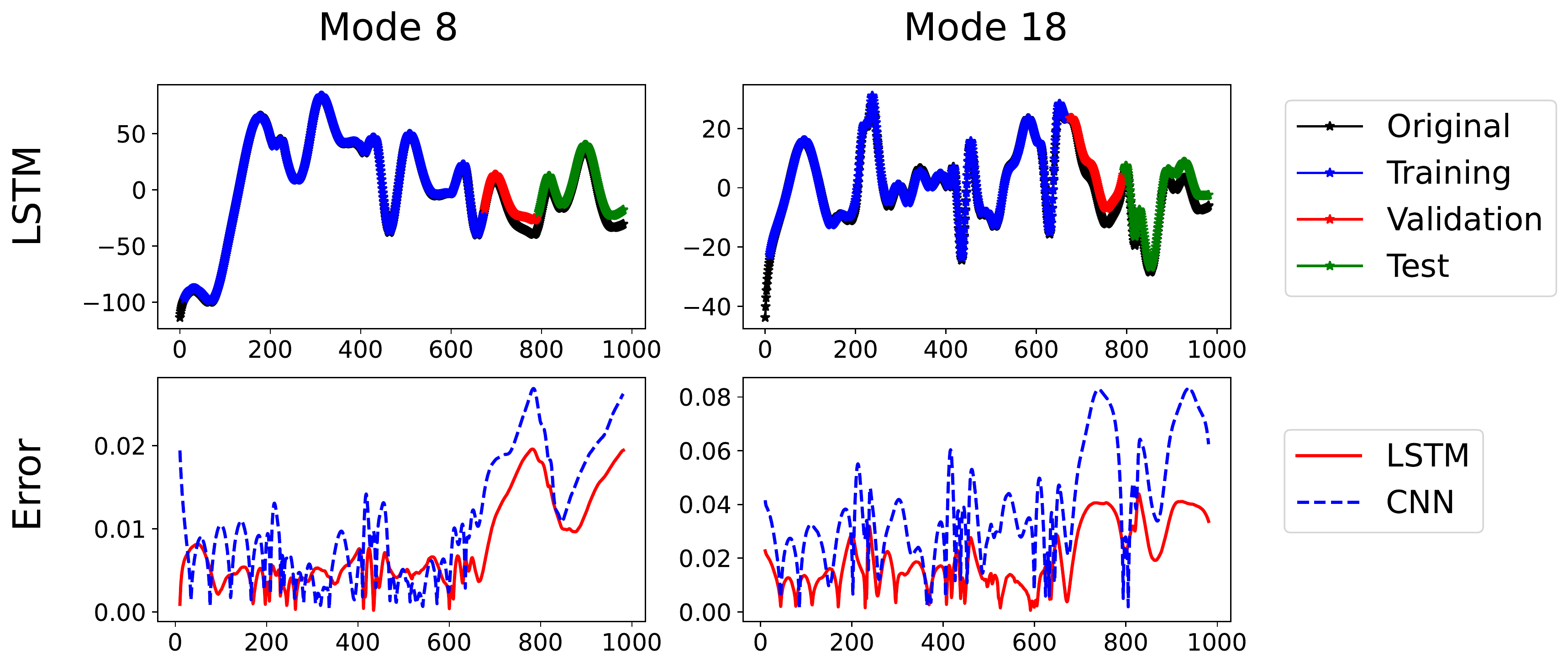}
	\caption{On top, prediction of the temporal modes $8$ and $18$ using the LSTM neural network. \textbf{Blue:} training set, \textbf{red:} validation set, \textbf{green:} test set and \textbf{black:} original mode. On the bottom, errors made in the prediction of the mentioned modes, for both LSTM and CNN models. }
	\label{fig:FigureA3}
\end{figure}

\end{document}